%% file: main.tex
\documentclass[11pt]{article}

\PassOptionsToPackage{table,dvipsnames}{xcolor}
\usepackage{xcolor}

\usepackage[utf8]{inputenc}

\usepackage{amsmath,amsthm,amssymb} 

\usepackage{graphicx}
\usepackage{tabularx}
\usepackage{booktabs}
\usepackage{multirow}
\usepackage{siunitx}
\usepackage{float}      
\usepackage{afterpage}  

\usepackage{algorithm}
\usepackage{algpseudocode}

\usepackage{enumitem}
\setlist[itemize]{leftmargin=-4ex,itemsep=1ex}

\usepackage{listings}

\usepackage{titlesec}
\usepackage{titling}
\usepackage{calligra,frcursive}
\usepackage[normalem]{ulem}

\usepackage[most]{tcolorbox}

\usepackage[font={small,sf},labelfont={small,sf,bf}]{caption}

\usepackage{fancyhdr}
\usepackage{hyperref}
\usepackage{cleveref}
\crefname{section}{§}{§§}

\definecolor{darkText}{HTML}{000000}
\definecolor{commentGreen}{HTML}{5C7E3A}
\definecolor{keywordBlue}{HTML}{0000FF}
\definecolor{functionPurple}{HTML}{7B3F99}
\definecolor{stringRed}{HTML}{D12F1B}
\definecolor{psiblue}{HTML}{3FA9F5}
\colorlet{psibluelight}{psiblue!20!white}

\lstdefinestyle{pythonlight}{
    backgroundcolor=,  
    basicstyle=\color{darkText}\ttfamily\small,
    commentstyle=\color{commentGreen}\itshape,
    keywordstyle=\color{keywordBlue}\bfseries,
    stringstyle=\color{stringRed},
    showstringspaces=false,
    breaklines=true,
    frame=none,
    numbers=none,
    language=Python,
    keywords={def, return},
    alsoletter={0123456789},
    emph={extract_optical_flow, copy, gaussian_bump, random_subset, psi, torch, cat, kl_divergence, idx_to_hw, argmax},
    emphstyle=\color{functionPurple},
}

\newcommand{\highlight}[1]{\colorbox{yellow!20}{\textbf{#1}}} 

\let\OLDthebibliography\thebibliography
\renewcommand\thebibliography[1]{%
  \OLDthebibliography{#1}%
  \setlength{\parskip}{0pt}%
  \setlength{\itemsep}{0pt plus 0.3ex}%
}

\title{{\LARGE \bfseries World Modeling with Probabilistic Structure Integration}}

\author{%
\href{https://neuroailab.stanford.edu/people.html}{Stanford NeuroAI Lab$^*$}
}

\date{\today}

\begin{document}

\maketitle
\thispagestyle{empty}

\input{sections/abstract}

\setcounter{tocdepth}{2}
\tableofcontents
\bigskip
\noindent$^*$Contributors to this work include: Klemen Kotar$^{\ast}$, Wanhee Lee$^{\ast}$, Rahul Venkatesh$^{\ast}$, Honglin Chen$^{\ast, \dagger}$, Daniel Bear$^{\dagger}$, Jared Watrous, Simon Kim, Khai Loong Aw, Lilian Naing Chen, Stefan Stojanov$^{\dagger}$, Kevin Feigelis$^{\dagger}$, Imran Thobani, Alex Durango, Khaled Jedoui, Atlas Kazemian, and Dan Yamins$^{\ast}$. (\ *=core contributor, $^{\dagger}$=alum)

\newpage

\input{sections/introduction}
\input{sections/prediction}
\input{sections/structure_extraction}

\input{sections/integration}

\input{sections/applications}
\input{sections/discussion}
\input{acknowledgements}

\bibliographystyle{unsrt}
\bibliography{refs}

\end{document}

%% file: sections/abstract.tex
\vspace{-10pt}
\noindent We present Probabilistic Structure Integration (\textbf{PSI}), a system for learning richly controllable and flexibly promptable world models from data. PSI consists of a three-step cycle. The first step, \textbf{P}robabilistic prediction, involves building a probabilistic graphical model $\Psi$ of the data, in the form of a random-access autoregressive sequence model. $\Psi$ supports a complete set of learned conditional distributions describing the dependence of any variables in the data on any others set of variables. In step 2, \textbf{S}tructure extraction, we show how to extract underlying low-dimensional properties in the data, corresponding to a diverse set of meaningful ``intermediate structures'', in a zero-shot fashion via causal inference on $\Psi$.  Step 3, \textbf{I}ntegration, completes the cycle by converting these structures into new token types that are then continually mixed back into the training diet as conditioning signals and prediction targets. Each such cycle augments the capabilities of $\Psi$, both allowing it to model the underlying data better, and creating new control handles -- akin to an LLM-like universal prompting language. We train an instance of $\Psi$ on 1.4 trillion tokens of internet video data; we use it to perform a variety of useful video prediction and understanding inferences; we extract state-of-the-art optical flow, self-supervised depth and object segmentation; and we use these structures to support a full cycle of predictive improvements.
\begin{center}
    \centering
    \captionsetup{type=figure}
    \captionsetup{labelfont=bf}
    \includegraphics[width=0.8\textwidth]{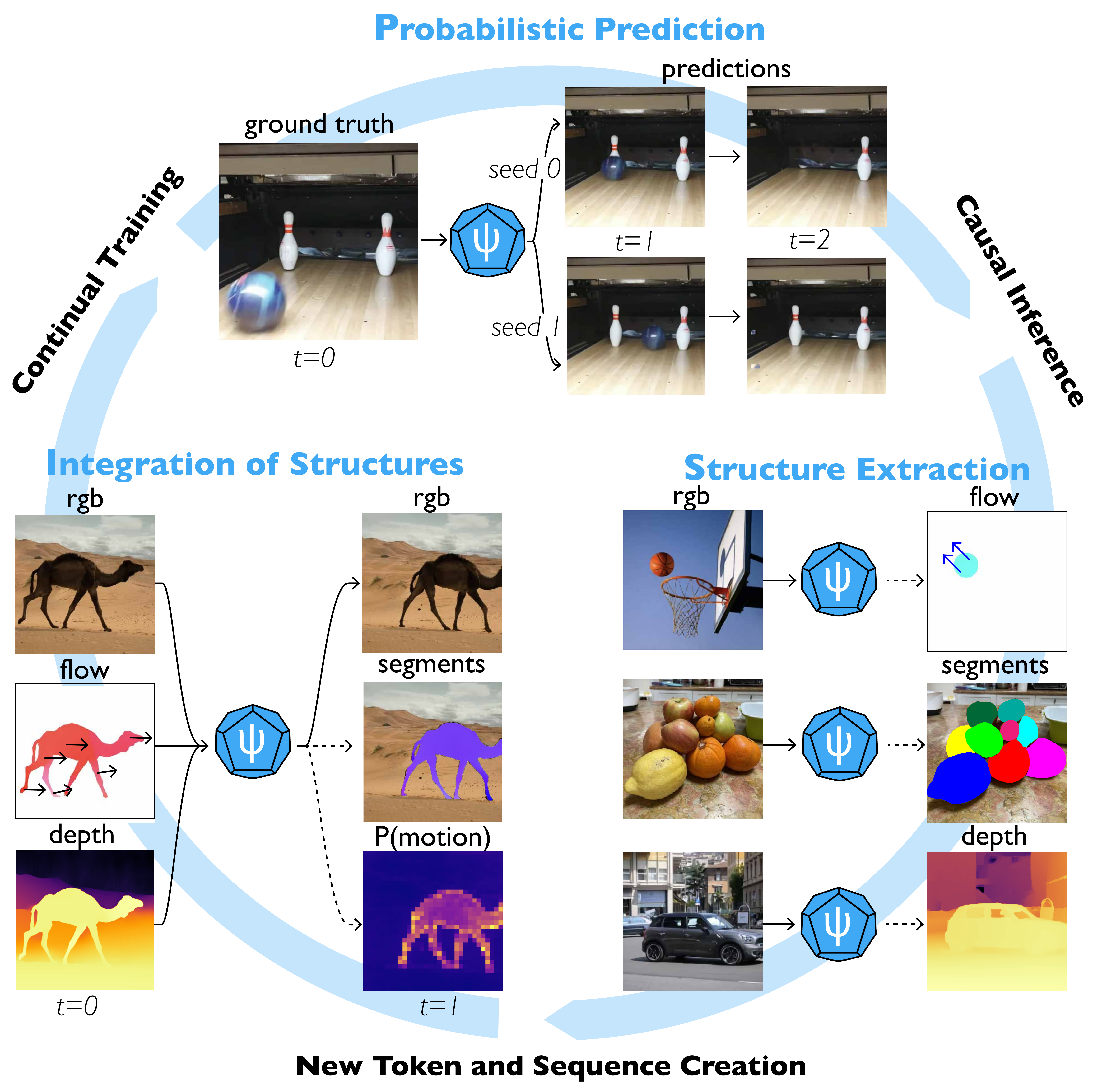}
    \caption{\textbf{Probabilistic Structure Integration.} We propose a new system for constructing self-improving world models consisting of a three-step cycle. First, a probabilistic predictor is trained on raw input data (e.g., RGB pixels). Second, the model is prompted to produce counterfactual generations, which isolate underlying structures in the data (e.g., optical flow, depth, segments). Third, those structures are integrated back into the predictor as both conditioning and prediction targets through continual training to improve its predictions and expand its capabilities.}
    \label{fig:teaser}
\end{center}

%% file: sections/introduction.tex
\section{Introduction}

The natural world is a complex system governed by a set of physical laws~\cite{newton1687principia}. Intelligent agents can exploit this property by forming models of the world to help reason about causal relationships and plan actions that achieve complex goals~\cite{hafner2020dreamer, ha2018worldmodels, Silver2017Predictron}. However, despite significant progress in learned simulators~\cite{bruce2024genie, yang2023learning} and generative AI~\cite{podell2023sdxl, blattmann2023stable}, today's world models have some key limitations, including a lack of \emph{rich controllability} and \emph{flexible query-ability}. Unlike pure language models, which can in their own domain be prompted in almost arbitrarily flexible (and enormously fun) ways, the ``API'' for interacting with the non-linguistic portions of world models --- e.g., the visual, motor, or memory components -- is often coarse and awkward, relying on discrete ``game-like'' action sets~\cite{bruce2024genie} or short text prompts~\cite{cosmos2024, blattmann2023stable} that very imperfectly capture the physical needs of real-world planning and control. In these settings, one cannot readily nudge an articulated limb, peel back an occluder, adjust the viewpoint, or manipulate specific objects -- precluding the richness of interaction that enables complex behavior. Foundation models in non-language domains are also awkward to query to expose their underlying world understanding, relying on representational learning with supervised probes to extract any given feature of interest. In language, a single model with a unified prompting interface excels at both controllability and query-ability; in vision, the absence of such an interface forces us to build separate models for generative and discriminative tasks.

In this work, we propose \textbf{Probabilistic Structure Integration (PSI)}, a generic process for building richly controllable world models with the right interfaces for true interactivity -- even when starting from low-level nonlinguistic raw input data.  PSI consists of three steps that form a virtuous cycle.  Step~1 involves the training of a \textbf{P}robabilistic predictor $\Psi$ that answers flexible visual queries at a local, low level of detail--close to raw data. Step~2 involves extracting \textbf{S}tructures from the predictor $\Psi$, corresponding to natural intermediate features for understanding the world -- effectively, new token types that transcend the low-level raw inputs.  Step~3 involves \textbf{I}ntegrating those new token types back into the predictor $\Psi$, using them as both new conditioning sources and prediction targets. The PSI process can be iterated, with each cycle enlarging the model’s control surfaces, enabling the construction of further new structures, and steadily increasing the predictive fidelity of the model itself.  A key feature of PSI is that the model interface remains the same throughout the process, with no need for expanded architectures of complex new types as the model self-enriches.   

The core mechanism of Step 1 is to have $\Psi$ predict a complete set of conditional distributions of local variables in the world, conditioned on arbitrary subsets of other such local variables. For example, in the case of natural video --- the main case we explore empirically in this work --- local variables correspond to (properly tokenized) spatiotemporal patches of RGB values.  Here, $\Psi$ learns to predict the conditional distribution of any patch in a visual clip, conditioned on an arbitrary subset of other such patches.  Because the world branches into multiple valid futures from any given observation, $\Psi$ represents the full conditional distribution in each prediction rather than a single best guess, so it can both sample plausible outcomes and compute high-level scene statistics, such as conditional entropy.  This interface supports a wide range of inference pathways about the scene: nudge a small area and see what else might change; reveal a few local observations and have the model synthesize coherent completions of the rest of the scene; or just perform unconditional predictions about the distributions over possible future states.  In other words, the $\Psi$ model is like a \uline{learned probabilistic graphical model} over patch variables, in which any observation can condition any prediction.  Learning a full probabilistic model of this kind may sound like a challenging problem: indeed, the key technical advance in Step 1 of PSI (see \S\ref{sec:prediction}) is formulating the $\Psi$ model so that it can be scalably learned from raw video data.  Specifically, we show how to construct $\Psi$ as a ``random‑access'' autoregressive transformer that learns stepwise conditional distributions on many random orderings of the same sparse visual data.  This formulation has the twin advantages: being extremely data efficient and as enabling us to repurpose the highly scalable modern LLM code infrastructure while avoiding task‑specific heads or architectural forks. 

Low-level variables can only go so far, however, in supporting complex world understanding and control goals. This is where the \textbf{S}tructure-extraction aspect of PSI comes in. Physically, structure extraction is based on the observation that interacting with the RGB-based $\Psi$ reveals natural groupings that exist in a scene (e.g., pixels that move together) and thereby exposes lower‑dimensional regularities in the underlying structure of reality. From the probabilistic graphical model view, we can intervene on any subset of local variables and measure induced changes elsewhere. By designing manipulations that elicit correlated responses across nodes, we expose axes of covariance and thereby uncover the latent structures that cause those correlations. The core technical insight (see \S\ref{sec:structure_extraction}) is that such structures --- sometimes called \emph{intermediate} representations or \emph{properties} --- arise in a \uline{zero-shot fashion as causal inferences in $\Psi$}, which in turn are comparisons between $\Psi$-predictions under factual and hypothetical interventions. Classical computer‑vision quantities are concrete instances of these low‑dimensional factors: \emph{optical flow} is the causal effect of appearance on future position~\cite{Horn1981Determining}, \emph{depth} arises from the causal relationship between viewpoint change and parallax~\cite{saxena2008make3d}, and \emph{segments} identify elements whose pixels share common fate under motion~\cite{shi2000normalized}. With a sufficiently controllable predictor, these properties can be extracted zero‑shot via counterfactual prompts. The idea of visual prompting was pioneered by Counterfactual World Modeling (CWM)~\cite{Bear2023CWM} with regression predictors; in PSI's \emph{distributional} setting, $\Psi$ produces sharp, coherent, multimodal outcomes that cleanly isolate causal effects where regression collapses them into blurry averages.

Intermediate structures can double as concise interfaces for describing and analyzing scenes, reducing complexity while aligning with physical organization. This observation motivates the third PSI step, where we close the loop by \textbf{I}ntegrating the extracted structures back into the predictor $\Psi$.  The core technical insight of this step is a \uline{very simple generic mechanism for new token-type integration}.  Essentially, this mechanism has three parts: (1) create the new tokens from outputs of step 2, using the same kind of tokenizer (and the same vocabulary) that works for RGB; (2) mix the new tokens into the middle of RGB token sequences, creating new mixed-type sequences for training to complement previous pure-type sequences; and (3) continue to train the same $\Psi$ model, on a mixture of the old and new sequences, from wherever it was previously left off training as if nothing had changed in the training diet.  Implementing this radically simple approach requires only a few key technical choices during implementation (as detailed in \S\ref{sec:integration}) --- the main empirical fact is that it works well, akin to mixing new synthetic data into an LLM’s diet mid‑stream~\cite{Wang2022SelfInstruct}.  The same unified model can then, for example, predict flow from RGB, generate RGB conditioned on flow, or compose these capabilities with depth and segments in arbitrary combinations.  Once integrated, each structure serves a dual role—as a conditioning signal and a prediction target—and, crucially, becomes a first‑class \emph{control surface} that enables richer interaction (e.g., specify flow to move an object, use depth to steer viewpoint, use segments to select and edit). This induces a self‑reinforcing cycle: improved predictions yield cleaner extractions; integrated structures expand and refine the control surfaces; and these, in turn, enable even better predictions and unlock new structures. Through this bootstrapping process, we progress from pixels to physics to broad scene understanding, supporting increasingly precise, local manipulation in the spirit of the rich representations envisioned by Marr~\cite{Marr1982Vision}.
\\

\noindent \textbf{Plan of this Paper:} \S\S~\ref{sec:prediction}---\ref{sec:integration} explore each of the steps of the PSI cycle in detail, and demonstrate the capabilities they unlock. \S\ref{sec:applications} highlights several proof-of-principle applications of the model to diverse downstream tasks. \S\ref{sec:discussion} contains a discussion relating PSI to a panoply of related topics in AI and cognitive science, and details some important open questions and next steps in the PSI framework. 

%% file: sections/prediction.tex

\section{Probabilistic Prediction: Richly Controllable World Models}
\label{sec:prediction}

We now turn to building $\Psi$, the controllable predictor at the heart of PSI.  The basic idea is to (1) break world data into a set of local variables and then (2) learn to predict the conditional probability distribution of each such variable, with the flexibility to condition on \emph{any} subset of these variables.  This idea is, in other words, to learn a \emph{probabilistic graphical model (PGM)}~\cite{koller2009probabilistic} of the variables characterizing the world.  With such a PGM in hand, a wide variety of inference pathways become possible, enabling fine-grained control.  In the case of video data, our main illustrative example here, the natural place to start for breaking the data up is to use spatiotemporal patches.  In this case, therefore, $\Psi$ begins as a probabilistic graphical model predicting the likelihood of observing any subset of pixel patches at specified locations within a video, conditioned on the observation of zero or more other such patches elsewhere.  

While the desire for a full probabilistic model of the world is clear, it might not at first be obvious how to obtain this from data in an efficient and scalable fashion. The main technical innovation of this section is showing how this can be done, using a technique we call Local Random-Access Sequence (LRAS) modeling.   

In what follows, we first describe $\Psi$'s probabilistic prediction objective more formally (Section~\ref{subsec:probabilistic_modeling}), and then introduce the LRAS technical formulation that enables learning $\Psi$ efficiently (Section~\ref{subsec:lras}).  We then illustrate how (Section~\ref{subsec:inference_pathways}) a specific instantiation of $\Psi$, trained on natural video data, can be used to perform a wide variety of inference paths, corresponding to useful controllability goals. Finally, we show how the probabilistic formulation supports a natural estimate of the entropy in scenes, and an iterative entropy-reducing ``uncertainty management'' process (Section~\ref{subsec:uncertainty_management}). 


\subsection{Defining the Probabilistic Relationships Between the Variables of the World}
\label{subsec:probabilistic_modeling}
The $\Psi$ model seeks to predict local distributions over plausible future world states as a function of previous conditioning.  To make this statement formal, we start with a data format that is described by a finite set of pointer locations $\mathcal{P}$ and a finite vocabulary $\mathcal{V}$ of content values.  A \emph{datum} $\mathbf{X}$ is any partially-defined function from $\mathcal{P}$ to $\mathcal{V}$, i.e., $\mathbf{X}: S \rightarrow \mathcal{V}$ where $S \subset \mathcal{P}$.  A \emph{pointer-structured data distribution} $\mathcal{D}$ is a distribution over such data.  A canonical example of this concept is RGB video clips, where the pointers are space-time locations $\mathcal{P} = \{(i, j, t) | i \in [0, H-1], j \in [0, W-1], t \in [0, T-1]\}$ that range over a two-dimensional discrete array and a time variable; the values $\mathcal{V}$ are 8-bit pixel intensities in the three standard color channels; and the distribution of valid data instances is described by the (very complicated) natural statistics of the real world.  
At one level, the pointer-structured data concept is somewhat vacuous -- it's hard to think of real-world data of interest that \emph{isn't} pointer-structured. However, since our ultimate goal is to create autoregressive models (\S\ref{subsec:lras}), it will turn out to be a useful formalism to handle the fact that many data types (e.g., vision) do not have a natural language-like uni-directional ordering. 

A key feature of pointer-structured data is that it is partially-defined in the pointer dimension, and potentially quite sparse. Partial definition expresses the concept of \emph{conditioning}, in that a predictor given such data can be asked to return responses that are consistent with the inputs.  For example, a valid datum might consist of a few patches of a single image, where natural ``future'' predictions complete the frame to a valid natural image.  Another example that we will use throughout this work is a two-frame clip, i.e., data of the form $\mathbf{X} = f_0 \cup \{x_0, \ldots, x_k\}$, where $f_0$ is a full video frame and $\{x_0, \ldots, x_k\}$ are a selection of patches from a nearby subsequent frame $f_1$. Natural predictions of this type of data are completions of the second frame that are consistent with the appearance of objects in $f_0$ and object motions sparsely revealed by $\{x_0, \ldots, x_k\}$.  

It is useful to think of the basic unit of pointer-structured data, the pointer-content pair $x = (p, v)$, as defining a single random variable in a probabilistic graphical model.  The collection of all such random variables over all the pointer locations is the set of random variables whose relationships will be modeled.  With these notions in mind, our core goal is to build a neural network for statistical completion of pointer-structured data. Formally, we are learning a function $\Psi$ that takes as input any valid datum $\mathbf{X}$, together with a pointer $\mathbf{p}$ not already specified in (the domain of) $\mathbf{X}$, and returns the conditional marginal probability in the joint distribution $\mathcal{D}$ of observing value $v$ at pointer $\mathbf{p}$, given conditioning $\mathbf{X}$.  In symbols,
\begin{equation}
\Psi: (\mathbf{X}, \mathbf{p} \notin \text{dom}(\mathbf{X})) \longmapsto \{\text{Pr}_{\mathcal{D}}[(\mathbf{p}, v) \text{ }|\text{ } \mathbf{X}] \text{ for } v \in \mathcal{V}\}.
\end{equation}
In other terms, $\Psi$ is a learned approximation to a probabilistic graphical model over the set of random variables corresponding to each pointer location.  

To make this concept more concrete, it's useful to think about what it means in the context of the two-frame video data described above.  In this case, $\Psi$ is a function from a single video still frame, together with a few patches of information about the next frame, and which returns a spatial map of distributions of values present in the next frame at the unspecified patches: 
\begin{center}
\includegraphics[width=0.7\linewidth]{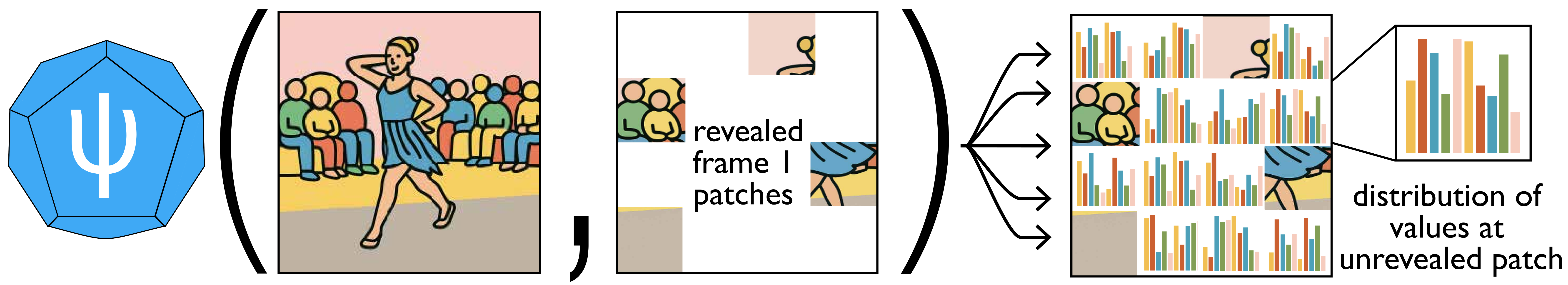}
\end{center}
In the above picture, the input to $\Psi$ is a single frame, as well as the few patches fully specified; in the output panel on the right, the model outputs distributions of predicted values at each of the unspecified locations, conditioned on the inputs being present.  Note that the sparse conditioning information about the second frame can, in this formulation, be of variable length.  An empty set of conditioning patches corresponds to an \emph{unconditional} prediction of the future frame, while as the size of the input patch set increases, further conditioning information is being provided, reducing the variability of the output distributions. With this idea in mind, it is convenient to define, for any datum $\mathbf{X}$ and subset of pointers $S \subset \mathcal{P} \setminus \text{dom}(\mathbf{X})$, the \emph{map of distributions}:

\begin{equation}
\Psi_S[\mathbf{X}] \equiv \{\Psi(\mathbf{X}, \mathbf{p}) \text{ for all } \mathbf{p} \in S\}.
\end{equation}
Since $\Psi$ predicts distributions of plausible values at any pointer location, it can be sampled to generate instances. Given an initial datum $\mathbf{X}$, we generate values at unspecified locations $\hat{\mathcal{P}} = \mathcal{P} \setminus \text{dom}(\mathbf{X})$ by iteratively selecting and sampling subsets. At each step $i$, we choose locations $S_i \subseteq \hat{\mathcal{P}} \setminus \left( \bigcup_{k < i} S_k \right)$ that have not been generated yet, then sample from the conditional distributions:
\begin{equation}
\mathbf{X}_i \sim \Psi_{S_i}[\mathbf{X}_{i-1}] \quad \text{where} \quad \mathbf{X}_0 = \mathbf{X}
\end{equation}
A major control knob for this process is the sampling order, described by the sequence of sets $S_i$ used during generation. Two basic ``pure'' sampling-order strategies are available:
\begin{itemize}
\item \textbf{Sequential sampling}: Each subset contains exactly one location ($|S_i| = 1$), requiring $n$ forward passes. This maximizes quality because each prediction attends to all previous patches.
\item \textbf{Parallel sampling}: A single subset contains all locations ($S_1 = \hat{\mathcal{P}}$), enabling one-pass generation but assuming conditional independence among patches.
\end{itemize}
Lying between sequential and parallel generation  are a wide variety of hybrid strategies---including generating patches in semantic or spatial groups---presenting a fundamental trade-off between quality and efficiency (see \S\ref{subsec:uncertainty_management}). 
Once the sampling order is decided, the ``usual'' sampling techniques from language models (e.g., temperature control or top-$k$,top-$p$ sampling) can also be used to control the diversity of predictions at each individual generation step. 

Throughout this paper, we use the notation:
\begin{equation}
\hat{\mathbf{X}} \sim \Psi[\mathbf{X}]
\end{equation}
to denote fully sequential generation at all unspecified locations with random decoding order.


\subsection{Local Random Access Sequence Modeling}
\label{subsec:lras}

\begin{figure}[!h]
    \centering
    \captionsetup{type=figure}
    \captionsetup{labelfont=bf}
    \includegraphics[width=0.98\textwidth]{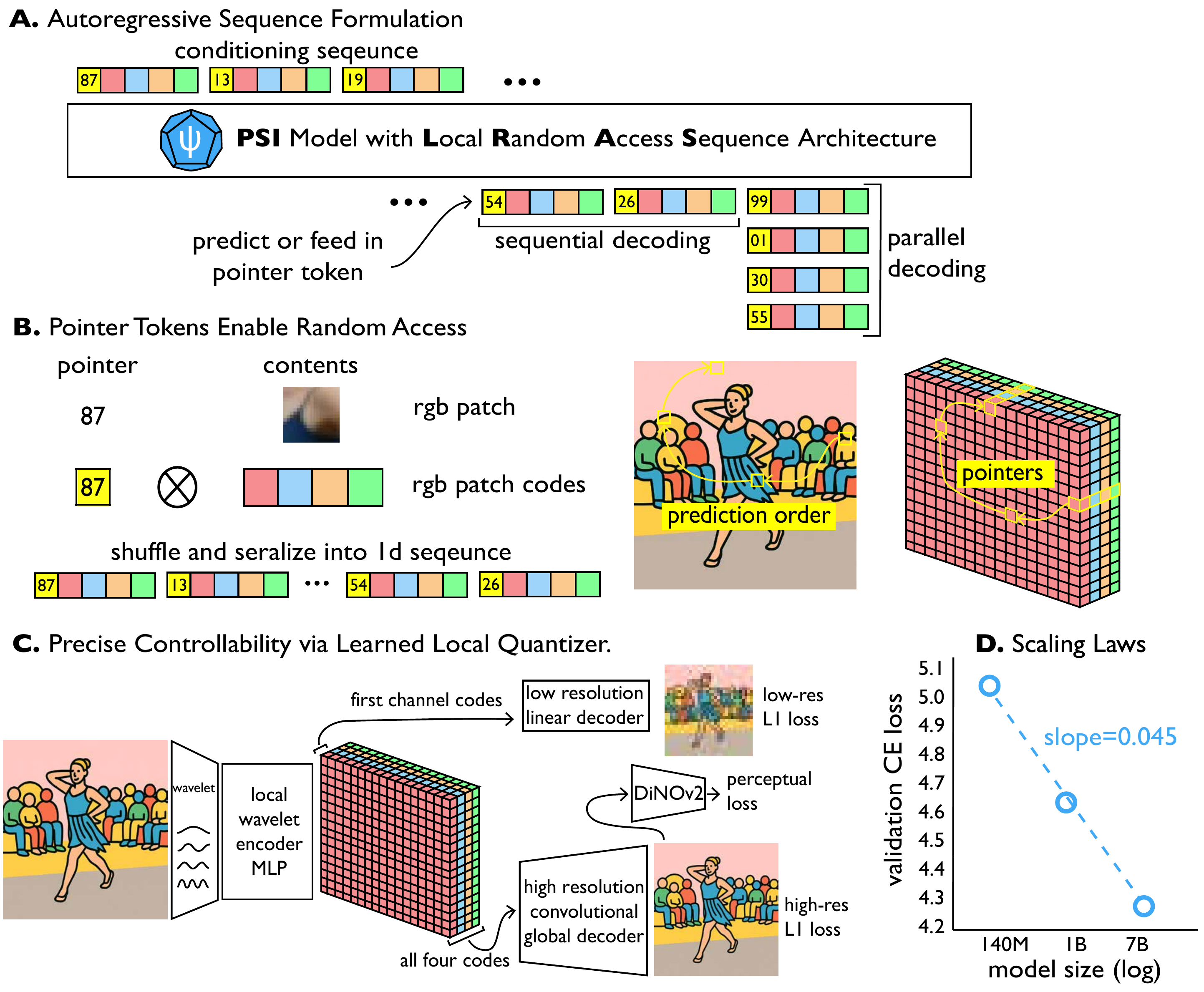}
    \caption{\textbf{LRAS Architecture.} \textbf{(A)}: Hierarchical Local Quantizer (HLQ) encodes each local patch into a sequence of local codes. The first code decodes to a low-resolution version of the input, while the rest provide details to upsample the reconstruction to the original resolution. \textbf{(B)}: HLQ codes are transformed into the Pointer-Content representation to allow for random access to the patches when encoding and decoding the image causally. \textbf{(C)}: Modeling the serialized image representation with an autoregressive transformer. Pointer tokens can be either predicted or seeded while decoding. This allows for any amount of sequential and parallel prediction from the same KV cache. \textbf{(D)}: Scaling Laws illustrating our model obtains predictably lower loss when given more parameters.}
    \label{fig:architecture}
\end{figure}

\noindent A key question we have not yet addressed is: how exactly can we formulate a neural-network learning objective to efficiently learn $\Psi$ as defined above. The key to this is the idea of the \emph{Local Random-Access Sequence} (LRAS) model~\cite{lee20253d}.  LRAS contains three key design decisions, which enable the implementation of the model described above:
\\

\noindent \textbf{Autoregressive Sequence Formulation.}
Because $\Psi$ seeks to make predictions with partial conditioning of variable size, an architecture in which conditioning data are represented by fixed-size tensors is inappropriate. Our core insight is that this problem can be solved by reformulating $\Psi$ as a sequence prediction problem (hence the ``S'' in its name), in which pointer values are simply woven into a sequence with one pointer $\mathbf{p}$ token before each value token $v$ --- so that each datum $\mathbf{X}$ is treated as a sequence $(\mathbf{p_0}, v_0, \mathbf{p_1}, v_1, \ldots)$.  With this in mind, the LRAS formulation reduces to the standard autoregressive framework:
\begin{equation}
\Psi(\mathbf{X}, \mathbf{p})
= \Psi(\mathbf{X} \!\circ\! \mathbf{p})
= \text{Pr}\bigl[v_k \mid \mathbf{p}_0, v_0, \mathbf{p}_1, v_1, \ldots \mathbf{p}_{k-1}, v_{k-1}, \mathbf{p} \bigr],
\end{equation}
and can be correspondingly trained with highly efficient sequence-to-sequence code. In this way, $\Psi$ architecturally resembles the most successful world models in modern AI (LLMs). In fact, $\Psi$ was designed to be optimized with the standard LLM pretraining procedure, utilizing the myriad of efficiency gains developed by that community to facilitate the production of ever larger models. 
\\

\noindent \textbf{Pointer Tokens Enable Random Access.}
\label{sec:pointer_content_data}
Instead of the standard GPT architecture we introduce Local Random-Access Sequence (LRAS) modeling, which adapts the causal autoregressive modeling paradigm for higher-dimensional data without a clear serialization order. Traditional GPT-style transformers predict sequences in a preset, hard-coded order. While this is a natural fit for one-dimensional data such as language, it becomes an unnecessary, and potentially harmful inductive bias when modeling higher-dimensional data. Most autoregressive image modeling approaches simply accept this bias, while we introduce a new token type --- the pointer --- which allows us to serialize the data in arbitrary order.

Pointer tokens enable us to package random-access traversals over high-dimensional data structures (such as images) into one-dimensional sequences of tokens for efficient causal pretraining, by interleaving pointer tokens among the content tokens, which represent the actual data, as illustrated in Figure \ref{fig:architecture}. During decoding, a \textit{pointer} token $i$ specifies the spatial index of the next patch to be emitted.  The model may either \emph{predict} a pointer (sampling the next region of interest) or \emph{consume} an externally supplied pointer, allowing the user to guide the generation pattern.

In addition to freeing us from the raster-order generation bias, pointer tokens allow us to condition our model on any subset of the image, learning complex multidirectional conditioning relationships in the data. They also allow for partial patch conditioning, and patch regeneration during inference. All of this functionality can still be simply expressed as a causal autoregressive sequence of tokens, and thus can be modeled and optimized as a standard LLM.
\\

\noindent \textbf{Precise Controllability via Learned Local Quantizer.}
\label{sec:hlq}
Most visual autoregressive models utilize popular off-the-shelf quantizers such as VQ-GAN~\cite{esser2021taming}, VQ-VAE~\cite{van2017neural}, or the recent Cosmos tokenizer~\cite{cosmos2024}. While such standard quantizers achieve strong compression ratios, they do not preserve locality of the patches within the token space, but rather encode the image in its entirety as a global code. While some locality is certainly present in the token representation, swapping any given token can modify the representations associated with patches on the other side of the image. While it is naturally more efficient to compress information globally, this comes at a cost of a less interpretable and less controllable latent (token) space.

Instead of compressing the whole frame into global codes, the LRAS architecture uses a Hierarchical Local Quantizer (HLQ): a convolutional autoencoder whose receptive field never crosses patch boundaries during encoding. Each patch is encoded into a sequence of four codes. The first code of each patch reconstructs a low-resolution preview, while the remaining codes add fine detail. No information from neighboring patches leaks into a code, preserving strict locality. This lets local downstream interventions---masking, overwriting, or re-ordering individual patches---behave predictably. Additionally, it makes the autoregressive modeling objective better aligned with natural-language modeling assumptions of token independence.
\\

\noindent \textbf{Key Training Details.}
We train an 80M-parameter HLQ on a combination of ImageNet and Open Images. We train a 7B-parameter $\Psi$ model on a dataset of 3 million RGB video clips, consisting of about 1.4 trillion video tokens. We train $\Psi$ with causal sequences of 2--4 frames, spanning up to 4 seconds of video. Mixed-precision training on 64 H100 GPUs at 65\% MFU yields 490 TFLOPS/device ($\sim$31 PFLOPS total) sustained.  

One critical ``detail'' of training is the use of the Warmup-Stable-Decay (\textbf{WSD}) learning rate schedule~\cite{Wen2024UnderstandingWL}.  In WSD, the learning rate is \textbf{W}armed-up briefly at the beginning of training, then held at a \textbf{S}table level for a period at as high a rate as possible while avoiding model divergence (with the S-phase length scaling with model size); and then, just before the model is desired for use, linearly \textbf{D}ecayed to zero just slowly enough to avoid divergence.  It has been shown that WSD achieves similar or better performance than cosine schedules of the same length. The use of WSD is particularly important here not because of superior performance, however, but because it enables continual training in the integration step of PSI (see \S\ref{sec:integration}).
\\

\noindent\textbf{Scaling Properties of LRAS.} A critical advantage of the LRAS architecture is its predictable scaling behavior, inheriting the well-established scaling laws of language models while extending them to visual domains. We trained $\Psi$ models from 100M to 7B parameters and observed consistent improvements in validation loss across three orders of magnitude (Fig.~\ref{fig:architecture}D). The reliable scaling without saturation even at 7B parameters suggests that further scaling would yield continued benefits, validating that LRAS successfully transfers the scaling properties of autoregressive language models to structured visual data.
\\

\noindent\textbf{LRAS Makes Probabilistic Graphical Models Tractable.} The LRAS formulation transforms the challenge of learning probabilistic graphical models at scale. The key insight is that our sequences represent traversals through the PGM---each pointer-content pair corresponds to visiting and observing a node in the graph. By modeling these traversals autoregressively, we approximate the full joint distribution through conditional factorization. The pointer mechanism ensures we can still query any conditional distribution $p(x_i | x_S)$ for arbitrary subsets $S$, but now through tractable sequential prediction rather than exponentially complex full inference. This recasts the seemingly intractable problem of learning a complete PGM over high-dimensional visual data as a standard GPT-style modeling problem.


\begin{figure}[H]
\centering
\captionsetup{type=figure}
\captionsetup{labelfont=bf}
\includegraphics[width=0.98\textwidth]{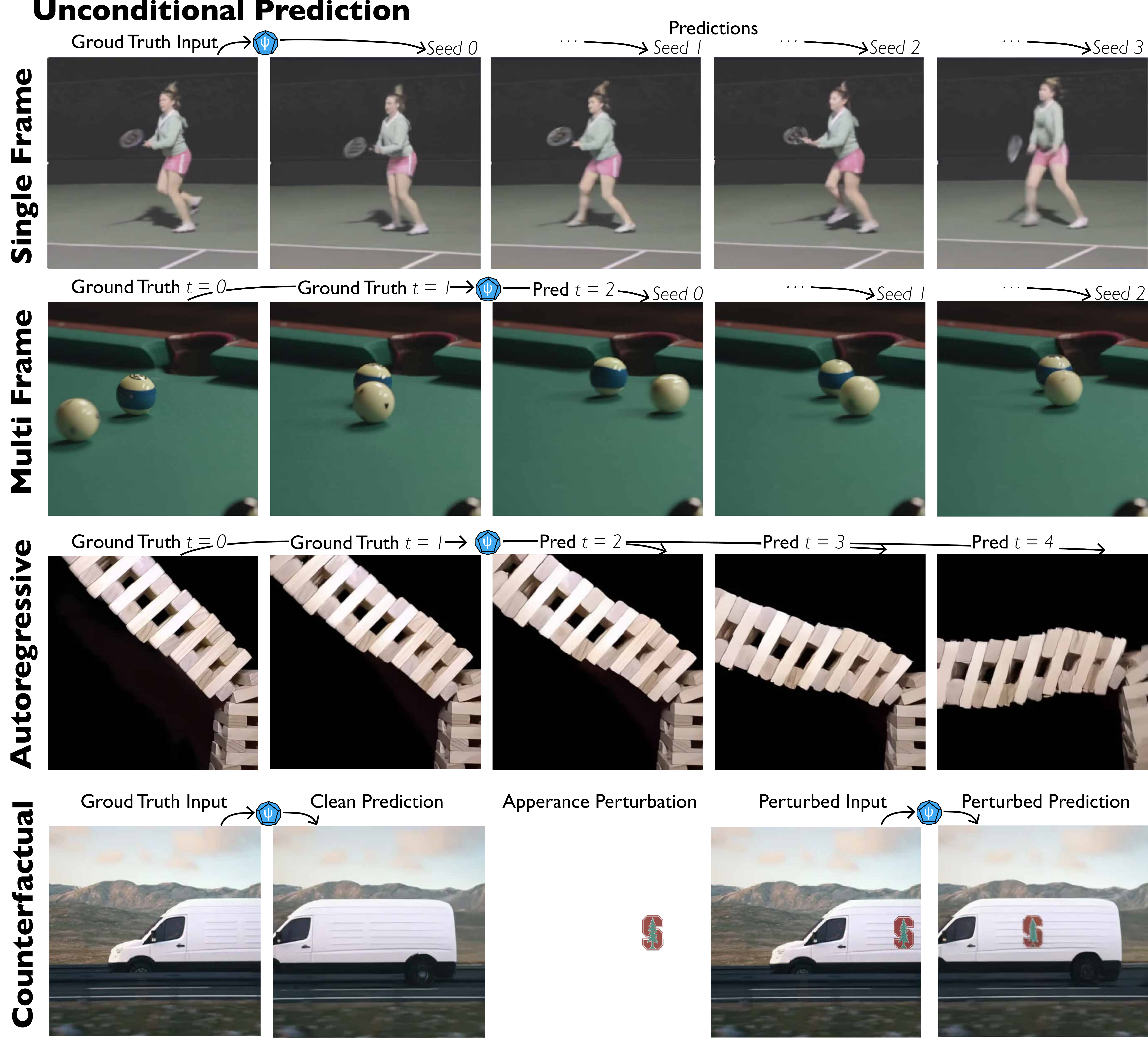}
\caption{\textbf{Unconditional Promptable Prediction.} Using the same unified model $\Psi$, single-frame unconditional prompting produces diverse plausible futures that reflect learned motion priors.}
\label{fig:unified_unconditional}
\vspace{-15pt}
\end{figure}

\subsection{Diverse Inference Pathways Confer Rich Controllability}
\label{subsec:inference_pathways}

\begin{figure}[H]
\centering
\captionsetup{type=figure}
\captionsetup{labelfont=bf}
\includegraphics[width=0.95\textwidth]{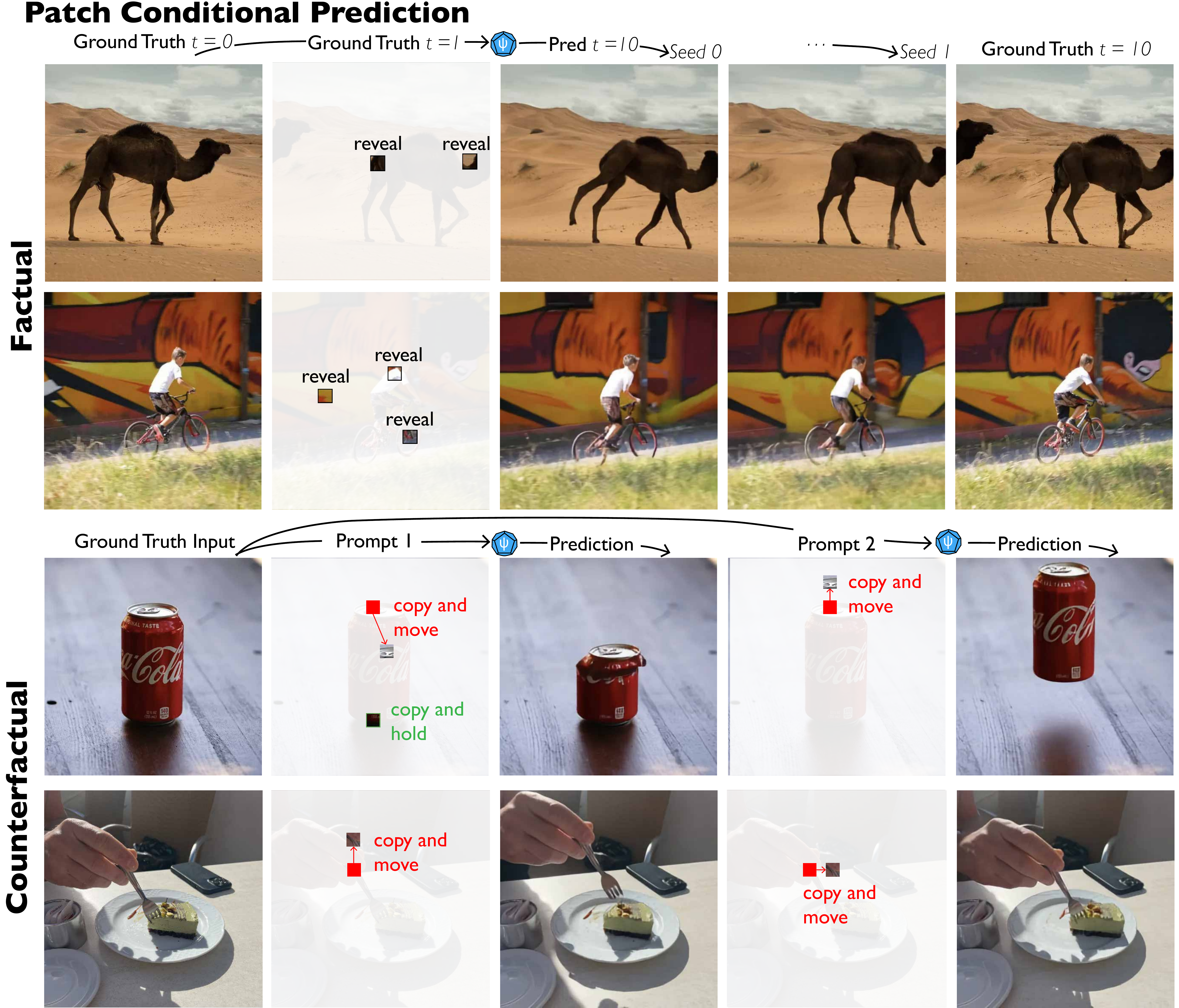}
\caption{\textbf{Patch-Conditional Promptable Prediction.} With sparse patches from the target frame as conditioning, $\Psi$ collapses uncertainty to a constrained completion; replacing patches with synthetic edits yields counterfactual manipulations while maintaining physical plausibility --- all with the same unified model.}
\label{fig:unified_patch_conditional}
\end{figure}

The LRAS architecture's flexible conditioning mechanism enables a rich spectrum of inference pathways through different prompting strategies. By varying which patches are provided as conditioning versus left for prediction, we can control $\Psi$ to perform diverse visual tasks---all without any task-specific training. Figures~\ref{fig:unified_unconditional} and~\ref{fig:unified_patch_conditional} illustrate some key inference pathways that emerge from this unified framework.\\

\noindent\textbf{Unconditional Prediction.} The simplest inference pathway predicts a future frame $f_1$ given only the current frame $f_0$:
\begin{equation}
f_{1} \sim \Psi\!\bigl[f_{0}\bigr]
\end{equation}
This represents the model's prior over how scenes evolve when no additional constraints are provided. The model must infer all motion from the static frame---guessing velocities, accelerations, and future trajectories. This leads to high diversity in samples as the model explores the full distribution of plausible futures. Providing multiple conditioning frames $f_{0}, ..., f_{t-1}$ constrains the trajectory by revealing motion patterns, though uncertainty about future dynamics remains.\\

\noindent\textbf{Patch-Conditional Prediction.} Since $\Psi$ treats all patches uniformly during training, we can condition generation on any subset of patches from the target frame. Let $f_1^P$ denote a sparse set of patches from frame 1:
\begin{equation}
f_{1}^{\bar P} \;\sim\; \Psi\!\bigl[f_{0},\,f_{1}^{P}\bigr]
\end{equation}
This dramatically constrains the prediction space---often fewer than 10\% of patches suffice to collapse generation to near-deterministic completion. The model leverages temporal factorization: appearance information comes from $f_0$ while motion cues come from the sparse patches in $f_1^P$.

Crucially, these conditioning patches need not be factual. By copying and displacing patches from $f_0$ to create hypothetical motion, we can prompt $\Psi$ to generate counterfactual scenarios. The model treats these synthetic patches as ground truth and synthesizes plausible completions---enabling precise control over object motion and scene dynamics.
\\

\noindent\textbf{Camera-Conditional Prediction.} When camera pose information is available, we can insert it directly into the sequence as additional tokens. Let $\mathbf{C}_{0\rightarrow 1}\!\in\!\mathrm{SE}(3)$ represent the relative camera transformation:
\begin{equation}
  f_{1} \;\sim\; \Psi\!\bigl[f_{0},\,\mathbf{C}_{0\rightarrow 1}\bigr]
\end{equation}
This conditions generation on known camera motion, enabling novel view synthesis. The model must still resolve ambiguities about scene scale and camera intrinsics, and must hallucinate previously occluded regions, leading to multimodal predictions that capture this inherent uncertainty.\\

\begin{figure}[H]
\centering
\captionsetup{type=figure}
\captionsetup{labelfont=bf}
\includegraphics[width=0.98\textwidth]{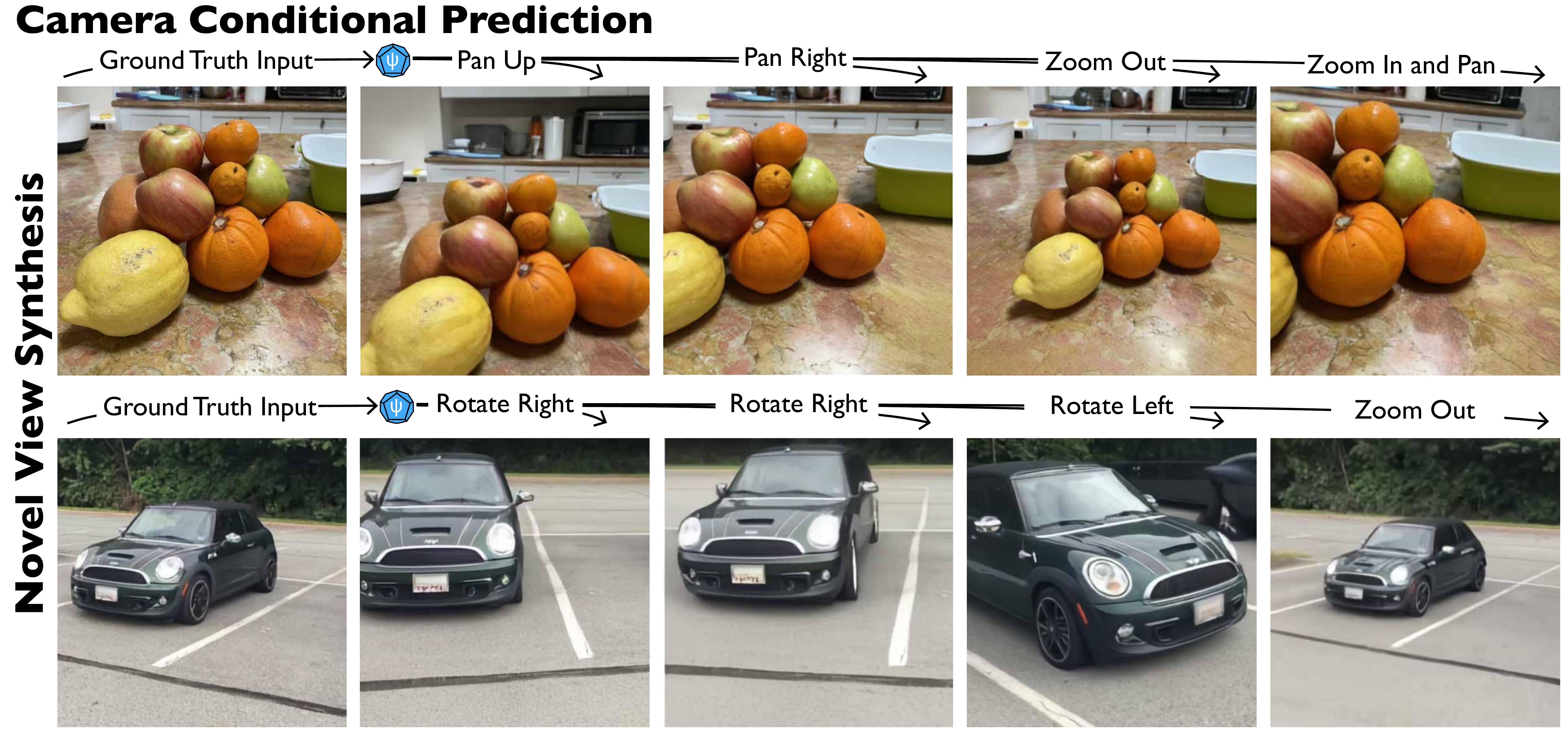}
\caption{\textbf{Camera-Conditional Prediction (Novel View Synthesis).} Given camera transformation parameters, $\Psi$ synthesizes new viewpoints that preserve scene consistency while hallucinating occluded regions, using the same unified model.}
\label{fig:unified_nvs}
\end{figure}

\noindent Put differently, these prompting patterns correspond to different conditional distributions over the visual data: unconditional prediction yields $p(f_{1}\mid f_{0})$, where the model must infer all future dynamics from a single frame; providing factual patches from the target frame conditions on partial observations $p(f_{1}^{\bar P}\mid f_{0},\,f_{1}^{P})$, constraining the completion to be consistent with the revealed patches; substituting synthetic patches implements a causal intervention $p(f_{1}^{\bar P}\mid f_{0},\,\mathrm{do}(f_{1}^{P}=\tilde f^{P}))$, forcing the model to generate completions consistent with the counterfactual evidence; and adding camera pose tokens conditions on additional observed variables $p(f_{1}\mid f_{0},\,\mathbf{C}_{0\rightarrow 1})$. In every case, the tokens supplied in the prompt are treated as observed evidence, and the model predicts the remaining tokens conditioned on this evidence.

These inference pathways are not separate capabilities but different prompts to the same unified model. The pointer-based architecture allows arbitrary combinations: we can simultaneously provide camera poses, factual patches, and counterfactual patches as conditioning. Each additional constraint reduces the entropy of the prediction, guiding $\Psi$ toward specific solutions within the vast space of plausible futures.


\subsection{Entropy and Uncertainty Management}
\label{subsec:uncertainty_management}

The pointer-based architecture of $\Psi$ enables flexible control over the generation process through its ability to decode patches either sequentially or in parallel. Sequential generation, where patches are sampled one at a time with each attending to all previous patches, provides maximal quality but requires multiple forward passes. Parallel generation enables single-pass inference but assumes conditional independence among patches. Between these extremes, we can generate patches in arbitrary groups, trading off quality for speed.

This flexibility becomes particularly powerful when combined with entropy analysis. By measuring the entropy of patch-wise logits during parallel prediction, we obtain a spatial map of the model's uncertainty:

\begin{equation}
  \mathcal{H} = 
  -\sum_{v \in \mathcal{V}} \Psi\bigl[f_{0}, ...\bigr] \log \Psi\bigl[f_{0}, ...\bigr]
\label{eq:psi_entropy}
\end{equation}

This entropy map $\mathcal{H}$ reveals which regions of the scene are most uncertain given current conditioning. As we sequentially generate or reveal patches, each new piece of information constrains the remaining predictions, progressively reducing uncertainty across the scene.

\begin{figure}[H]
    \centering
    \captionsetup{type=figure}
    \captionsetup{labelfont=bf}
    \includegraphics[width=0.85\textwidth]{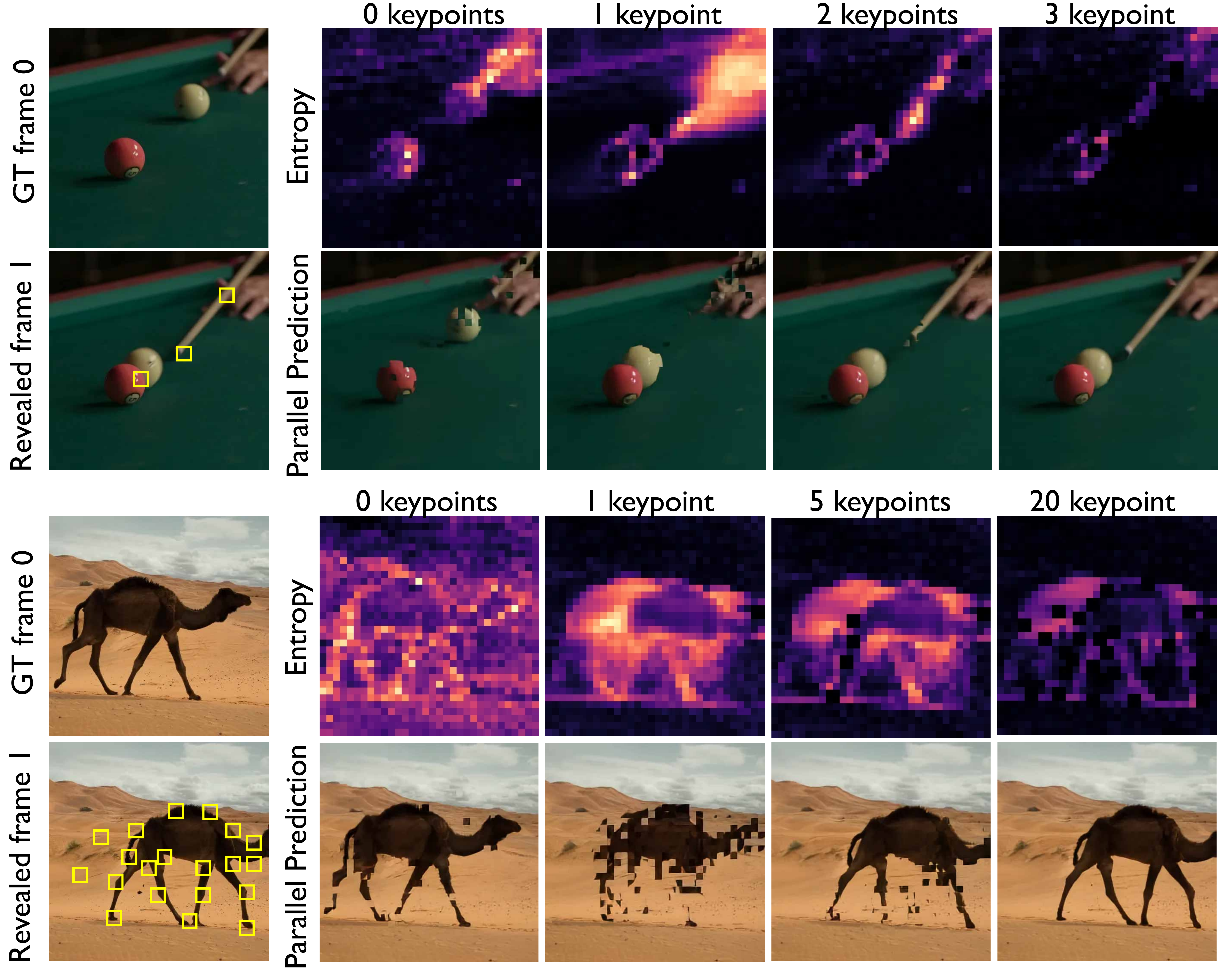}
    \caption{\textbf{Progressive Uncertainty Reduction Through Sequential Conditioning.} We perform a series of \textbf{parallel} predictions at each step to measure the model's \textbf{certainty} during generation. We reveal the patches that yield the highest entropy at each step to optimally resolve the scene uncertainty. Each column demonstrates how additional conditioning patches (yellow squares) progressively constrain the model's belief about the state of the world, and improve the parallel prediction. \textbf{Note that sequential rollouts will produce coherent, sharp generations with any amount of conditioning patches revealed, at the cost of more compute.}}
    \label{fig:uncertainty_management}
\end{figure}

Figure~\ref{fig:uncertainty_management} illustrates this process on a scene with a camel in motion. Initially, entropy concentrates on the animal's limbs and edges---regions with highest motion variability. As patches are revealed (yellow squares), the model updates its beliefs: first recognizing that the camel has moved, then resolving ambiguity about its new position. The rightmost column shows how parallel predictions at each stage become progressively more deterministic, converging from diverse samples to near-identical outputs as uncertainty collapses.

This unified view of sequential generation and uncertainty has important implications. For time-critical applications like robotic control, we can use entropy to adaptively choose between sequential and parallel generation---using fast parallel decoding when entropy is low, but switching to sequential generation in high-uncertainty regions. The entropy map also identifies which patches most efficiently reduce global uncertainty, enabling principled strategies for active perception and sparse conditioning.

The ability to quantify and manage uncertainty while flexibly controlling generation speed makes $\Psi$ practical for diverse applications, from real-time robot planning where speed matters more than perfect quality, to high-fidelity video generation where sequential consistency is paramount.

%% file: sections/structure_extraction.tex

\section{Structure Extraction: Prompts as Causal Inference}
\label{sec:structure_extraction}

Section \ref{sec:prediction} shows how to produce a learnable PGM for natural videos, where the variables are local spatiotemporal patches, leading to a powerful generative model.  However, it is natural to want to transcend patch-level variables so as to be able to create world-model controls at higher levels of abstraction --- e.g., motions and objects rather than RGB pixels.  Such control surfaces naturally manifest as \emph{intermediate (visual) structures} --- that is, visual quantities that humans can estimate from raw RGB data but which in most typical computer vision contexts are thought of as ``supervised labels'' --- e.g., optical flow, depth, object segmentation boundaries, and so on.   If such quantities were readily available, they would be natural control conditioners --- it would, for instance, be easier and more effective to create object motion hypotheticals by directly placing a patch of non-zero optical flow somewhere on the object rather than by the kind of naive pixel translation implemented in Section~\ref{sec:prediction}. 


\subsection{Intermediate Structures Are Zero-Shot Causal Prompts}
\label{sec:intermediate_structures_are_causal_probes}

But where should these intermediate quantities come from?  Of course, it might be possible to use third-party supervised extractors to estimate them as needed.  However, it turns out that a more conceptually satisfying and robustly performant approach is available.  A true world model can be prompted to perform tasks it was not explicitly trained for. In Section~\ref{sec:prediction} we explored various atomic conditioning prompts that allow us to control the generation of the model.  Now we will show how we can compose several of those prompts to extract intermediate structures in a zero-shot fashion from $\Psi$.  The basic concept is that:
\begin{center}
\begin{tcolorbox}[width=.75\textwidth, ams align*]
\text{Properties of interest } &=  \text{ Intermediate structures}\\
&= \text{ Structured prompts comparing factual and}\\&\quad\quad\text{hypothetical/counterfactual predictions from $\Psi$}\\
&= \text{ Causal inferences from $\Psi$}\\
&= \text{ Generators of new types of tokens.}
\end{tcolorbox}
\end{center}
The first and second equalities above are essentially the idea advanced in an earlier work on Counterfactual World Modeling~\cite{Bear2023CWM}, but here we apply (and improve) this concept using the LRAS model.  The third equality is essentially the main observation of the theory of probabilistic graphical models~\cite{Pearl2009Causality}, of which $\Psi$ is (as noted above) an instance.  The last equality will be exploited in the subsequent section to create improved, more richly controllable, versions of $\Psi$.  

This overall formulation also mirrors the paradigm shift from probing to prompting present in language models (and see \ref{subsec:probe_to_prompt}). Rather than training task-specific probes for each visual property, causal prompts serve as the visual analog to language prompts—structured counterfactuals that query scene properties zero-shot, without supervision.  Each extraction from $\Psi$ is obtained by holding most of the scene fixed, perturbing only the factor of interest --- in probabilistic graphical model terms, this is the equivalent of performing one or more $\texttt{do()}$ operator(s) on the PGM encoded by $\Psi$ --- and measuring the model's response.  The result is a unified mechanism for structure extraction that is steered entirely through prompt engineering.

Compared to the regression-based Counterfactual World Model (CWM) \cite{Bear2023CWM}, which pioneered the process of zero-shot quantity extraction with counterfactual prompts, $\Psi$ is a distributional generative model, and thus, offers two key advances. Firstly, the generated predictions are trained to match real-world distributions, not just their means, producing sharp predictions in response to hypothetical prompting, which makes it easier to precisely isolate the intermediate structure in question, for example, sharply segmenting the boundaries of an object. Secondly, $\Psi$ is probabilistic, allowing for the resolution of ambiguous situations where prompts have multiple discrete and valid answers. This enables $\Psi$ to capture complex definitions of intermediates such as objects, which can have fuzzy definitions.  Repeatedly, we will find that being able to perform ``crisp hallucinations'' --- that is, correct hypotheticals --- is key for making effective causal inferences --- that is, for good structure extraction.   

What follows are three sections that illustrate causal inference programs to create three key intermediate visual structures: \textit{optical flow}, \textit{object segments}, and \textit{depth} using $\Psi$.  These results are also discussed in separate standalone papers exploring each topic in greater depth: Kim et al.~\cite{kim2025taming} for optical flow (also see~\cite{stojanov2025self}), Venkatesh et al.~\cite{spelke_net} for object segments, and Lee et al.~\cite{lee20253d} for depth (which also illustrates novel view synthesis and complex object edits).


\subsection{Optical Flow via Tracer Counterfactuals}
\label{subsec:kl_tracing}

\textit{Causal Inference Definition:} When performing next frame prediction, the value of any given pixel in $f_1$ is ``caused'' by the appearance of pixels in $f_0$, which constrains the prediction. What translates the values between the two frames is the motion of the pixels --- otherwise known as an optical flow field. Therefore, we can expose this flow field by counterfactually perturbing the pixel values of $f_0$ and observing the impact on $f_1$, since the former causes the latter. By marking a dot in $f_0$ and recording where that dot moves to in $f_1$, we reveal the causal structure of motion.

The key advance of $\Psi$ over the regression-based Counterfactual World Model (CWM) \cite{Bear2023CWM} lies in its distributional generative nature. Where CWM produces blurry mean predictions, $\Psi$ generates sharp, coherent samples. This distinction is crucial for optical flow, where correspondence can be inherently ambiguous---a texture patch might plausibly match multiple locations, or an occluded region might have several valid disocclusion patterns. In such cases, the world admits multiple sharp, discrete futures, not a single blurry average. The mean of plausible solutions is almost never a valid solution itself. While these sharp predictions would cleanly propagate counterfactual tracers through the scene if decoded, $\Psi$'s distributional formulation offers something even more powerful: we can directly measure how perturbations affect the predicted distributions without ever decoding to RGB.

Following the Counterfactual World Modeling approach introduced by Bear et al. \cite{Bear2023CWM}, we extract optical flow by measuring the effect of a slight appearance perturbation on the next frame prediction. Specifically, we add a small Gaussian bump $\boldsymbol{\delta}$ to patch $i$ in frame $f_0$, creating $\tilde{f}_{0}^{(i)} = f_0 + \boldsymbol{\delta}\,\mathbf{1}_{\text{patch}=i}$. We then query $\Psi$ with both the original and perturbed frames, conditioning on a random subset $\mathcal{R}$ of revealed patches from $f_1$. In CWM, the difference is computed as an L2 distance between mean predictions, while $\Psi$, with its distributional nature, allows us to capture a richer metric --- the KL divergence between the predicted distributions. The patch whose distribution shifts most under perturbation identifies where the tracer reappears:

\begin{equation}
  j^{\star}(i)
  \;=\;
  \underset{j \in \mathcal{P}_{f_{1}}\setminus\mathcal{R}}{\arg\max}\;
  \mathrm{KL}\!\Bigl(
      \Psi\!\bigl[\tilde{f}_{0}^{(i)},\,f_{1}^{\mathcal{R}}\bigr]
      ,
      \Psi\!\bigl[f_{0},\,f_{1}^{\mathcal{R}}\bigr]
  \Bigr)
  \label{eq:kl_tracer_masked}
\end{equation}

\noindent The displacement vector $\mathbf{u}_{i}= j^{\star}(i) - i$ yields the optical flow estimate. This distributional approach naturally handles multimodality, producing diverse yet individually coherent flow predictions rather than attempting to average incompatible motion hypotheses.

\begin{figure}[H]
    \centering
    \captionsetup{type=figure}
    \captionsetup{labelfont=bf}
    \includegraphics[width=0.9\textwidth]{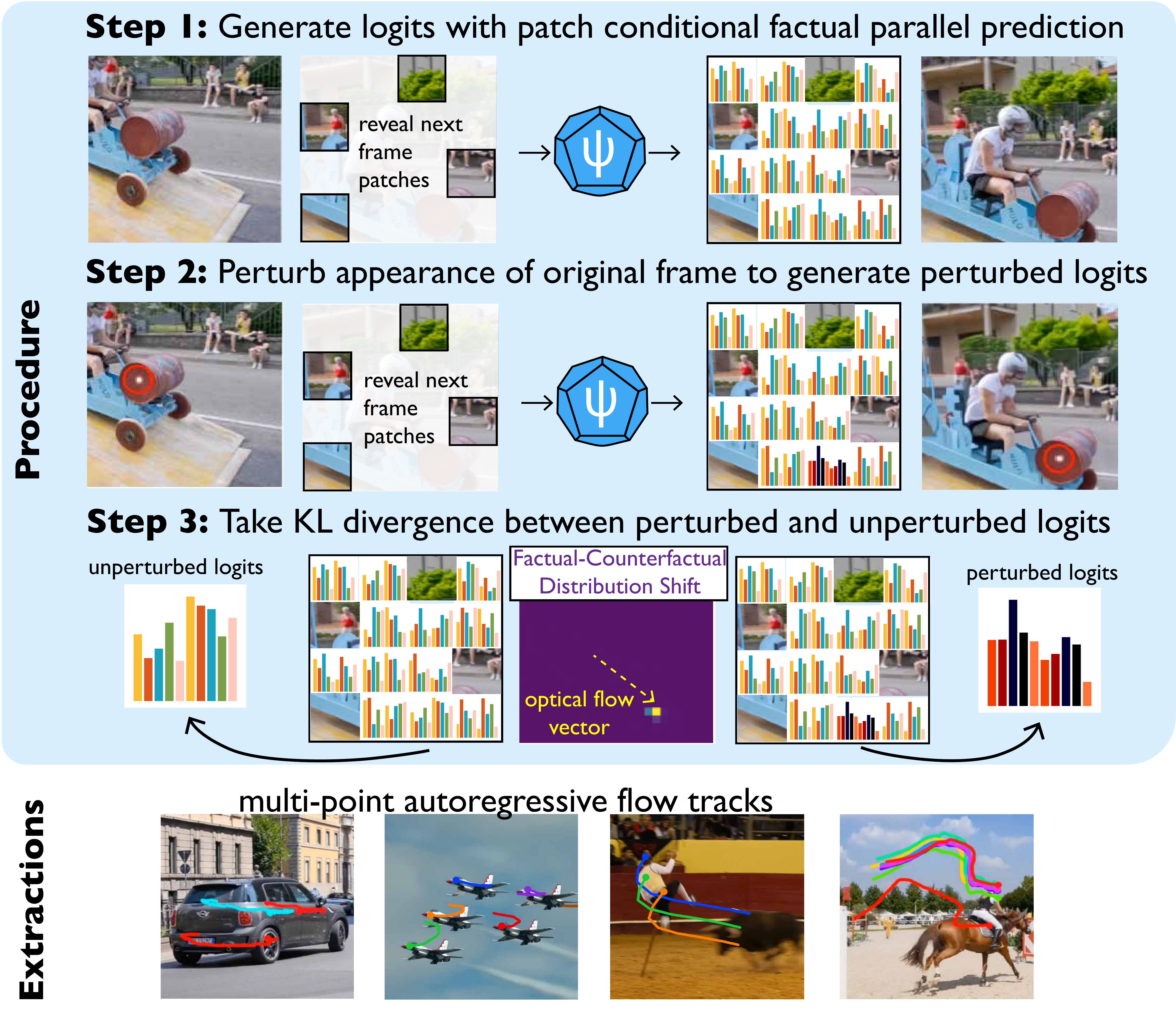}
    \caption{\textbf{Optical Flow via KL Tracing.} \emph{Procedure (top):} 
    \textbf{Step 1} — run patch‑conditional \emph{parallel} prediction to obtain per‑patch logits; 
    \textbf{Step 2} — add a tiny dot to $f_0$ to form a perturbed input and recompute logits; 
    \textbf{Step 3} — take the KL divergence between perturbed and unperturbed predicted distributions to locate the correspondence in $f_1$, yielding a flow vector. 
    \emph{Extractions (bottom):} flow tracks across diverse scenes produced by repeating the probe at several locations.}
    \label{fig:flow_extraction}
    \vspace{0pt}
\end{figure}

\begin{table}[H]
\renewcommand{\arraystretch}{1.2}
\centering
\tiny
\resizebox{\linewidth}{!}{%
\begin{tabular}{p{3mm}lccccccccccc}
\multirow{2}{*}{} &
\multirow{2}{*}{\textbf{Method}} &
\multicolumn{4}{c}{\textbf{DAVIS}} &
\multicolumn{4}{c}{\textbf{Kubric}} \\
\cmidrule(lr){3-6}
\cmidrule(lr){7-10}
& &
AJ~$\uparrow$ & AD~$\downarrow$ & $<\!\delta^x_\text{avg}\!$~$\uparrow$ & OA~$\uparrow$ &
AJ~$\uparrow$ & AD~$\downarrow$ & $<\!\delta^x_\text{avg}\!$~$\uparrow$ & OA~$\uparrow$ \\
\midrule
S
& RAFT~\cite{teed_raft_2020}
& 41.77 & 25.33 & 54.37 & 66.40 & 71.93 & 5.60 & 82.15 & 88.54  \\

& SEA-RAFT~\cite{wang_sea-raft_2024}
& 43.41 & 20.18 & 58.69 & 66.34
& \highlight{75.06} & 6.54 & \highlight{84.63} & \highlight{89.50} \\

\midrule
W
& Doduo~\cite{jiang_doduo_2023}
& 23.34 & \underline{13.41} & \underline{48.50} & 47.91 & 54.98 & \underline{5.31} & 72.20 & 73.56  \\
\midrule
U

& SMURF~\cite{stone_smurf_2021}
& \underline{30.64} & 27.28 & 44.18 & 59.15
& \textbf{65.81} & 6.81 & \underline{80.57} & \textbf{87.91} \\

& CWM \cite{bear_unifying_2023, venkatesh_counterfactual_2023}
& 15.00 & 23.53 & 26.30 & \textbf{76.63}
& 28.77 & 11.64 & 41.63 & 84.93\\

& PSI (ours)
& \textbf{44.16} & \textbf{11.18} & \textbf{65.20} & \underline{74.58} & \underline{65.49} & \textbf{5.06} & \textbf{81.66} & \underline{87.63}  \\
\end{tabular}
}
\caption{\textbf{TAP-Vid First: quantitative results on DAVIS and Kubric.}
Tracking starts when a point first appears and continues to the video end, thus involving large frame gaps.
$\Psi$ outperforms two-frame baselines. “S" = supervised, “W" = weakly supervised, “U" = unsupervised.}
\vspace{0pt}
\label{tab:flow_results}
\end{table}

Figure \ref{fig:flow_extraction} illustrates this advantage. The top panel shows the KL-tracing procedure, computing divergence between clean and perturbed patch-conditioned predictions. The bottom panel demonstrates flow traces extracted from real-world dynamic scenes, tracking points across multiple frames using only the initial frame as conditioning. This zero-shot approach achieves state-of-the-art results on TAP-DAVIS and TAP-Kubric benchmarks (Table \ref{tab:flow_results}), surpassing both task-specific and world-model-based methods. For detailed analysis and quantification, see~\cite{kim2025taming}.


\subsection{Object Segments via Motion Hypotheticals}
\label{sec:rgb_segmentation}

\textit{Causal Inference Definition:} When observing scene dynamics, the coherent motion of pixels across frames is ``caused'' by their membership in objects that move as unified entities. Groups of pixels move together because they belong to the same physical object—a latent structure that constrains their collective motion. Therefore, we can expose these object boundaries by hypothetically perturbing the motion of a small patch and observing which other pixels are compelled to move in concert, revealing the causal structure of object membership through their shared fate.

Following the approach pioneered by Bear et al. \cite{Bear2023CWM}, we extract object segments by forcing $\Psi$ to resolve motion hypotheticals. We copy a patch from $f_0$ and place it at a shifted position in $f_1$, creating the hypothetical prompt $f_{1}^{\boldsymbol{\Delta p}} \sim \Psi\!\bigl[f_{0}, i \rightarrow i + \boldsymbol{\Delta p}\bigr]$. Since the model understands that objects move coherently, it displaces all pixels belonging to the same object to maintain physical plausibility. We then compute optical flow between $f_0$ and the hypothetical prediction using SEA-RAFT \cite{wang_sea-raft_2024}, with the induced motion field revealing object boundaries. This transforms implicit object understanding into explicit segment tokens without task-specific supervision.\footnote{The results of this section mirror a more detailed presentation in a standalone paper on the use of the PSI technique to perform object segmentation~\cite{spelke_net}.}

The critical advance of $\Psi$ over CWM lies in its distributional modeling of both object motion and background synthesis. While CWM produces mean predictions that can blur object boundaries, $\Psi$ generates sharp hypothetical motions---diverse yet individually coherent samples that capture the multimodal nature of object behavior. This is particularly important for non-rigid objects that can deform and move in complex ways---a person's limbs might swing, a flag might flutter, or a tree's branches might sway. By sampling multiple hypothetical motions with different seeds, we can explore the space of plausible object behaviors and determine pixel membership through their consistent co-movement across samples. When an object moves hypothetically, $\Psi$ must synthesize the previously hidden background with sharp detail---a task with multiple valid solutions that the model represents through its predictive distribution rather than collapsing to an averaged blur. Moreover, $\Psi$'s multimodal predictions naturally handle the inherent ambiguity in object definitions: are a teacup and its plate one object or several? Are a rider and bicycle separate entities or a functional unit? Different samples might segment the scene differently based on which motion hypothesis is realized, aligning with the fuzzy, context-dependent nature of object perception where the same scene admits multiple valid interpretations.

\begin{figure}[H]
    \centering
    \captionsetup{type=figure}
    \captionsetup{labelfont=bf}
    \includegraphics[width=0.8\textwidth]{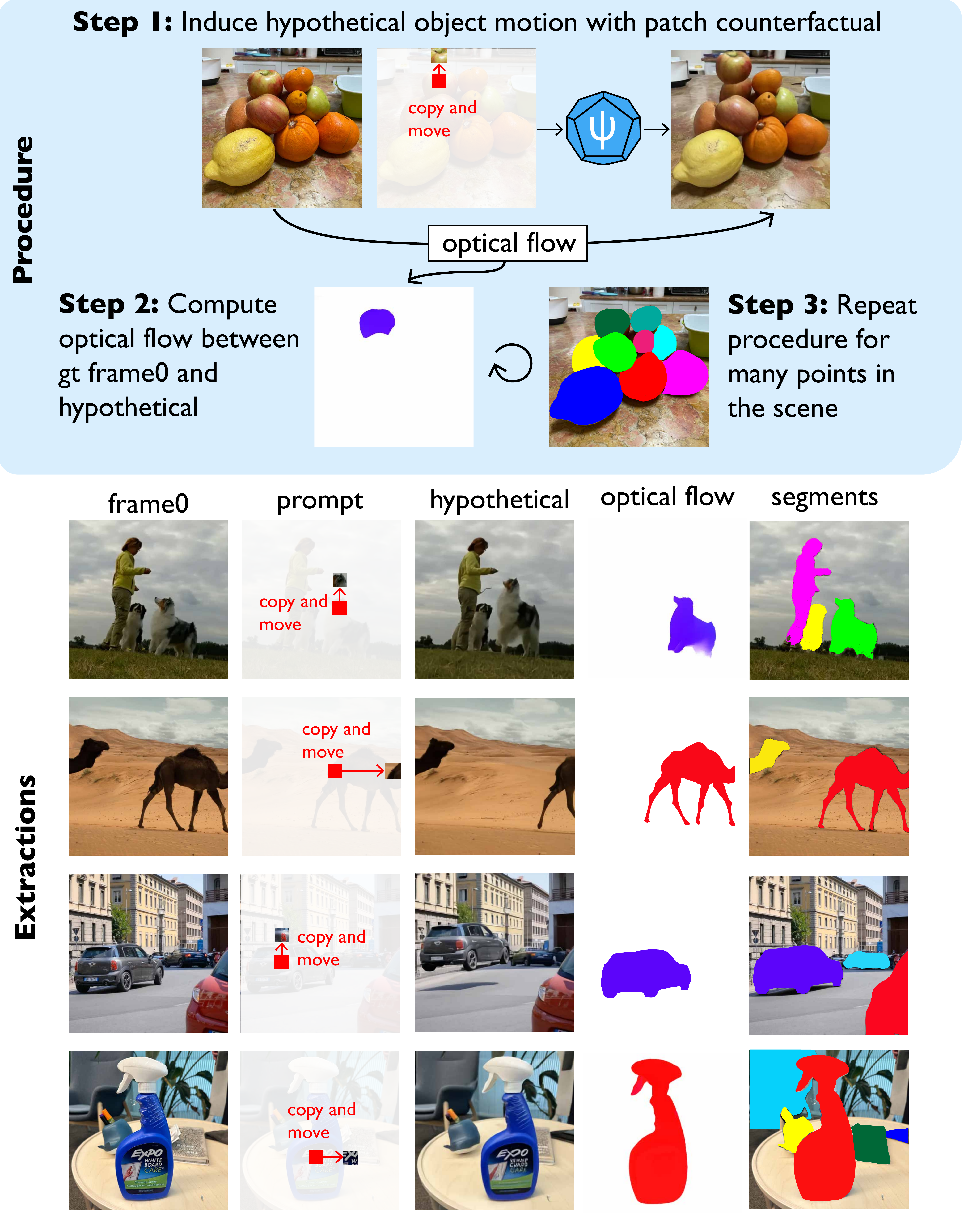}
    \caption{\textbf{Object Segments from Motion Hypotheticals.} \emph{Procedure (top):} 
    \textbf{Step 1} — induce object motion by copying a small patch in $f_0$ and moving it to a new location (counterfactual prompt); 
    \textbf{Step 2} — generate the hypothetical frame and compute optical flow between $f_0$ and the hypothetical; 
    \textbf{Step 3} — repeat over seed locations to cover multiple instances and convert coherent, induced motion into segments. 
    \emph{Extractions (bottom):} for each scene: frame $f_0$, prompt, hypothetical completion, induced optical flow, and the resulting instance segments.}
    \label{fig:rgb_segmentation}
    \vspace{-10pt}
\end{figure}

Figure \ref{fig:rgb_segmentation} showcases motion hypothetical prompts and their resulting segmentations. The model's sharp predictions enable clean object isolation --- note how it appropriately generates occluded portions and respects depth ordering. For dynamic scenes, extreme counterfactuals (e.g., lifting a car upward) work particularly well, as they violate typical motion priors and force the model to isolate the object's causal influence. Quantitative performance is described in Section~\ref{subsec:better_extractions}. More sophisticated algorithms leveraging this causal principle are discussed in \cite{spelke_net}.

\subsection{Depth via Viewpoint Hypotheticals}
\label{sec:rgb_depth_extraction}

\textit{Causal Inference Definition:} When a camera translates through 3D space, the differential motion of pixels in the image plane is caused by their depth --- the distance from the camera to the scene point. Near objects exhibit larger displacements than distant ones, a phenomenon known as motion parallax. Therefore, we can expose this depth structure by hypothetically perturbing the camera viewpoint and measuring the resulting parallax field, since depth is the latent cause that governs the magnitude of induced motion under viewpoint changes.

We extract depth by prompting $\Psi$ to generate novel viewpoints. We append camera translation tokens $\mathbf{C}_{\Delta} \in \mathrm{SE}(3)$ representing a pure in-plane shift of magnitude $b$, yielding $f_{1}^{(\Delta)} \sim \Psi\!\bigl[f_{0},\,\mathbf{C} \rightarrow \mathbf{C} + \mathbf{C}_{\Delta}\bigr]$. This creates an artificial stereo pair from which we compute optical flow using SEA-RAFT \cite{wang_sea-raft_2024}. For each patch $j$, the horizontal disparity $d_j = |u_j|$ relates to depth through $z_j = b/d_j$, revealing the 3D structure without any depth supervision.\footnote{The ideas of this section echo a more detailed presentation of results using the PSI technique to achieve 3D scene understanding, novel view synthesis, object editing, and unsupervised depth extraction in a standalone paper~\cite{lee20253d}.}

\begin{figure}[H]
    \centering
    \captionsetup{type=figure}
    \captionsetup{labelfont=bf}
    \includegraphics[width=0.8\textwidth]{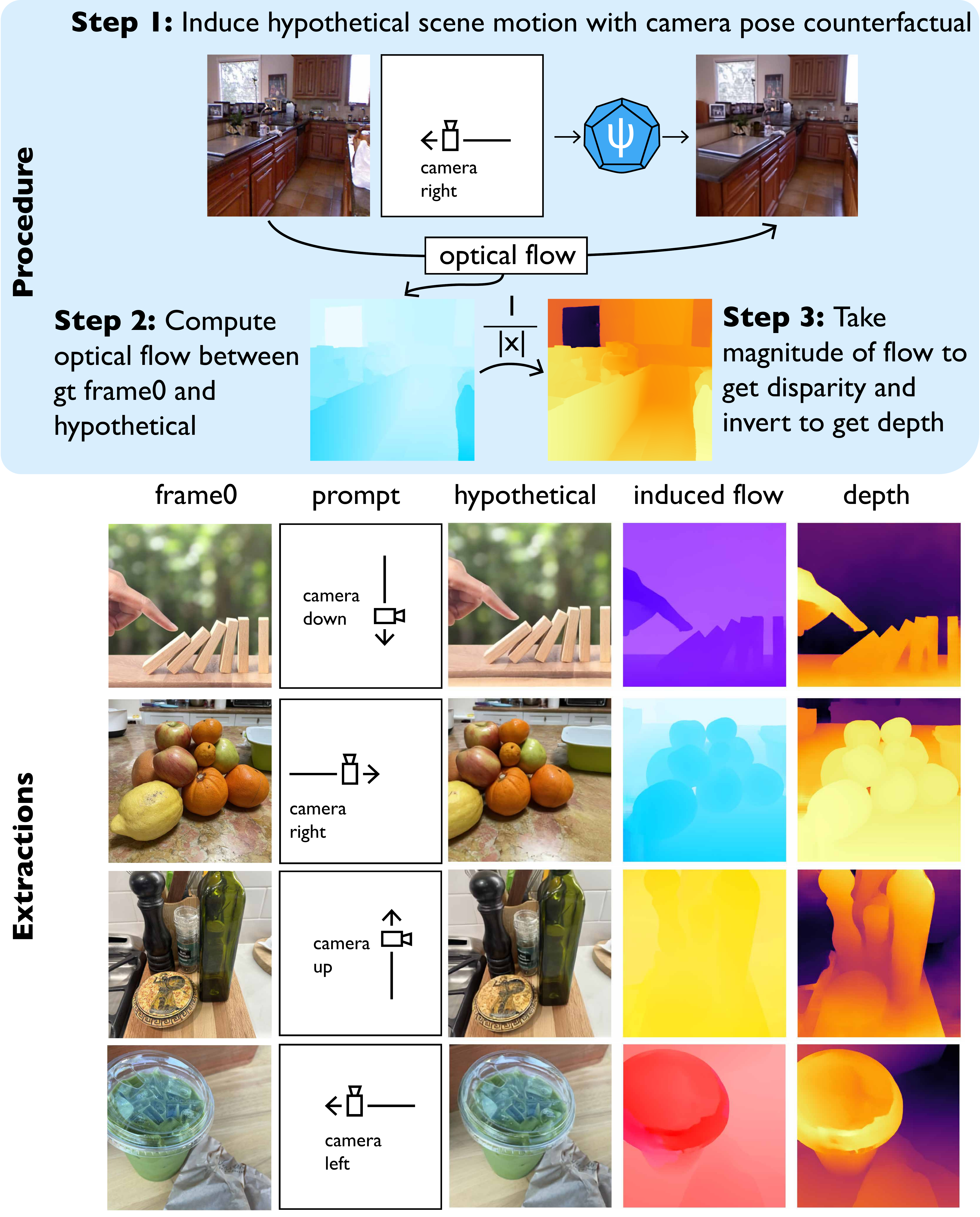}
    \caption{\textbf{Depth from Viewpoint Hypotheticals.} \emph{Procedure (top):} 
    \textbf{Step 1} — apply a camera‑pose counterfactual (e.g., translate right/left/up/down) and generate the hypothetical view; 
    \textbf{Step 2} — compute optical flow between $f_0$ and the hypothetical; 
    \textbf{Step 3} — take the flow magnitude as disparity and invert to obtain (unscaled) depth. 
    \emph{Extractions (bottom):} per scene, we show the prompt, hypothetical, estimated flow, and resulting depth, illustrating consistent geometry across varied motions and baselines.}
    \label{fig:rgb_depth}
\end{figure}

The advantage of $\Psi$'s distributional modeling is particularly pronounced for depth extraction, where good hypotheticals --- sharp novel views from camera motion --- are essential for accurate parallax computation. While blurry predictions coming from regression models (like CWM) make precise stereo correspondences difficult, $\Psi$ generates crisp, geometrically consistent views that enable clean disparity measurements. This sharpness is crucial at occlusion boundaries where $\Psi$ must synthesize previously hidden regions with geometric accuracy. Moreover, $\Psi$'s distributional nature elegantly handles depth ambiguities in textureless regions or through transparent surfaces, where multiple depth interpretations may be equally valid. Rather than collapsing to a single estimate or meaningless blur, $\Psi$ represents multimodal depth distributions that capture inherent uncertainty. For dynamic scenes, $\Psi$'s multimodal predictions are particularly valuable---sometimes the model predicts object motion along with camera motion, sometimes it does not. By averaging depth estimates across multiple samples, we can isolate the consistent depth signal from variable object dynamics, yielding robust depth maps even in moving scenes.

Note that this method builds upon optical flow as a fundamental extraction—the causal chain proceeds from depth to parallax to flow, highlighting how these intermediate representations compose. Figure~\ref{fig:rgb_depth} visualizes depth maps produced solely via viewpoint hypotheticals, demonstrating how $\Psi$'s sharp predictions enable fine geometric detail. The model maintains consistency across different baseline translations, from subtle vertical shifts that freeze dynamic scenes to larger horizontal movements that reveal complex depth structure. Quantitative performance is described in Section~\ref{subsec:better_extractions}.


\subsection{Extracted Structures Define New Token Types}
\label{subsec:new_token_types}

The structures extracted through causal inference are not merely outputs---they correspond to latent variables in the scene’s probabilistic graphical model (PGM). Flow links past appearance to future position; depth mediates camera motion and parallax; segments capture object membership and coherent motion. Viewed this way, each structure is a natural token type: a discrete, local handle the model can read or write alongside RGB. As in language (characters to words to phrases), a multi-level visual vocabulary lets $\Psi$ represent appearance, motion, geometry, and semantics within one sequence. Crucially, these tokens are bidirectional interfaces, serving as both conditioning inputs and prediction targets without task-specific heads or new vocabularies. 

This perspective turns extraction into a bridge to integration. Once structures are tokenized, we can interleave them with RGB in a single autoregressive sequence to build a more expressive and powerful model. Section~\ref{sec:integration} operationalizes this bridge with a universal mechanism that mixes new token types into the sequence and continues training.

%% file: sections/integration.tex

\section{Integrating Structure: 
Ratcheting Prompting and Control}
\label{sec:integration}

Having demonstrated how $\Psi$ extracts fundamental visual structures through causal inference prompts, we now show how these structures can be integrated back into the model as new token types, dramatically expanding its prompting vocabulary. This integration transforms extracted intermediates from mere outputs into first-class citizens of the model's representational space, enabling richer and more precise interactions.


\subsection{A Universal Mechanism for Structure Integration}

Predictors that could directly observe and manipulate the new token types arising from structure extraction would be of great use.  For example, if one built LRAS predictors combining flow tokens with the original RGB tokens in various combinations, it could unlock entirely new prediction ``APIs'', where the new tokens can be used \emph{both} as inputs (e.g.,\ conditioning signals) and outputs (e.g.,\ prediction targets):
\begin{equation} \nonumber
o \sim \Psi_1[f_0] \quad\quad\quad o \sim \Psi_2[f_0, c] \quad\quad\quad f_1 \sim \Psi_3[f_0, o].
\end{equation}

In the above, $\Psi_1$ predicts the flow field $o$ following $f_0$, learning the distribution of potential motions consistent with the input image $f_0$. $\Psi_2$ conditions on camera pose information $c$ alongside $f_0$, disentangling global camera motion from object motion in the predicted flow field. $\Psi_3$ uses both $f_0$ and $o$ as conditioning to produce $f_1$---creating a powerful motion renderer that warps pixels according to the precise specifications of $o$, while synthesizing occluded regions. 

What if one now wanted to integrate depth maps $d$ in the same way? New predictors could capture these relationships:
\begin{align*}
&d \sim \Psi_4[f_0] \quad\quad\quad d \sim \Psi_5[f_0, c] \quad\quad\quad f_1 \sim \Psi_6[f_0, d],\\
&d \sim \Psi_{7}[f_0,o] \quad\quad d \sim \Psi_{8}[f_0, o, c] \quad\quad f_1 \sim \Psi_{9}[f_0, d, o].
\end{align*}
Similar to the case with flow, $\Psi_4, \Psi_5, \Psi_6$ predict and condition on depth, while $\Psi_7, \Psi_8, \Psi_9$ combine flow and depth in various prediction and conditioning relationships. Each of these relationships is useful and complementary. However, the proliferation of useful predictor configurations needed to define these extended APIs is inconvenient. Must we train a zoo of specialized models to capture these manifold useful prediction problems? Many existing approaches would require architectural modifications for each new task---diffusion models need specialized conditioning mechanisms, regression models require new output heads. Even approaches that add extra channels or embedding heads suffer from parameter explosion and fail to scale computation with the increased information. Most critically, such architectures miss the key property we seek: establishing directional causal relationships between old and new tokens. A bidirectional transformer or encoder-decoder cannot specify whether flow causes RGB or RGB causes flow---but in our autoregressive formulation, the position of tokens in the sequence defines their causal ordering, making each new structure a proper node in the learned probabilistic graphical model.

It turns out that the design of LRAS enables a very simple but powerful unified solution to this problem.  This three-step mechanism integrates any extracted structure into a unified model, requiring no architectural changes as new structures are added. This recipe transforms new structures into new tokens in the sequence, enabling seamless expansion of the model's capabilities through simple continual training.
\\

\noindent\textbf{Step 1. Tokenize the data and assign new pointers.}
For spatial intermediates such as optical flow and depth, we use the same HLQ architecture (Section~\ref{sec:hlq}) employed for RGB to produce 2D token grids that preserve locality and allow precise, patch-level control. For object- or scene-level intermediates such as segments, we adopt variable-length tokenizations that escape the \textit{tyranny of the grid}~\cite{dieleman2025latents}, assigning as many tokens as needed per object rather than forcing a fixed patching. After tokenization, we add new \emph{pointer addresses} for these structures by extending the pointer set $\mathcal{P}$ while keeping the \emph{value vocabulary} $\mathcal{V}$ fixed and simply reuse the token indices for different modalities, differentiated by the pointers (Section~\ref{sec:pointer_content_data}). Thus, a previously RGB-only model can accept flow or depth tokens just by addressing them at new pointer locations---no new heads, no architecture changes, and, crucially, no vocabulary expansion that would cause parameter explosion. This design allows us to integrate an arbitrary number of new structures while keeping the model size stable, as each new modality only requires new pointer addresses rather than expanding the embedding matrices or output layers.
\\

\begin{figure}[H]
    \centering
    \vspace{-10pt}
    \captionsetup{type=figure}
    \captionsetup{labelfont=bf}
    \includegraphics[width=0.98\textwidth]{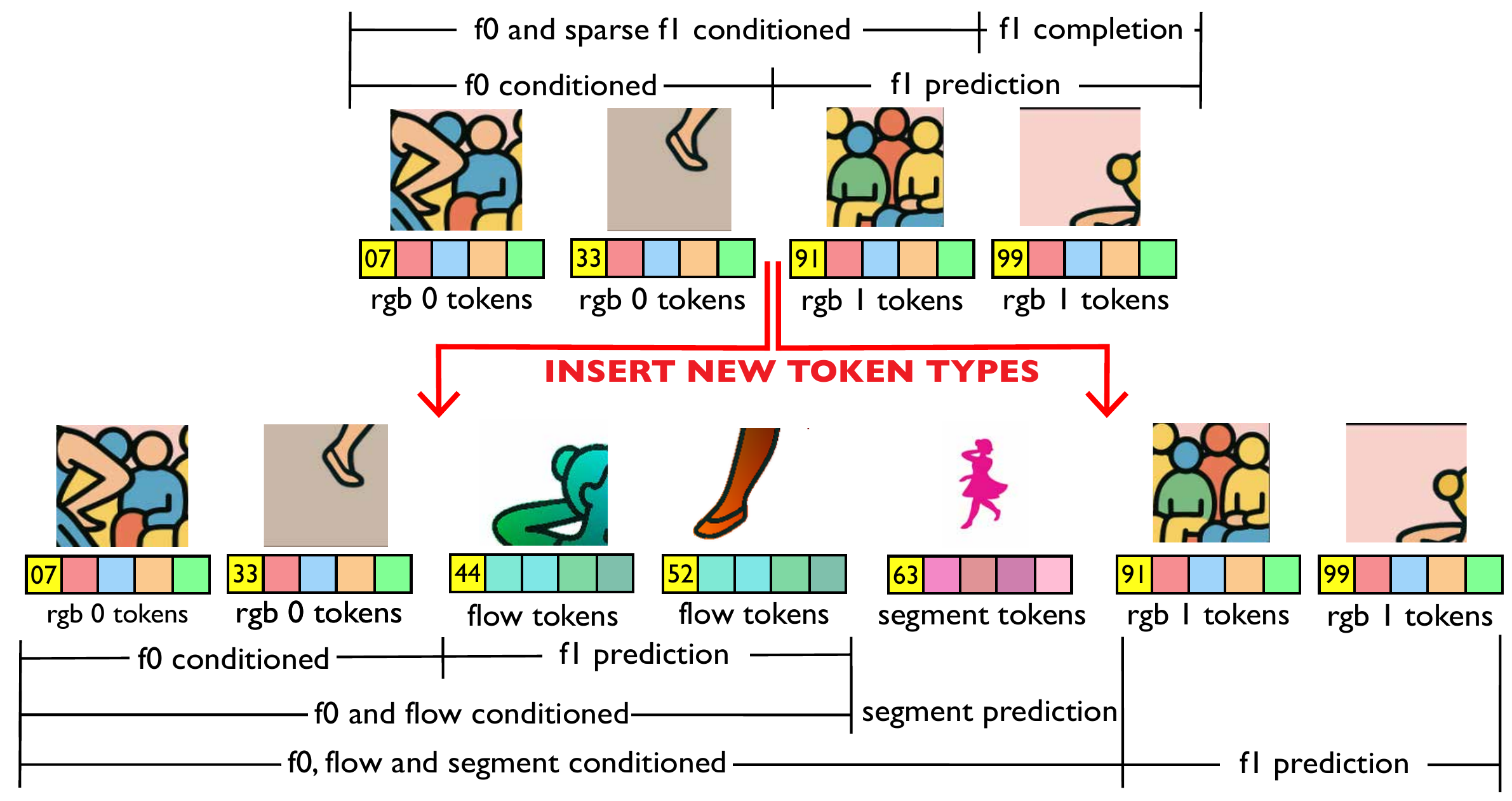}
    \caption{\textbf{Mixing New Tokens Into Sequences (Integration Step 2).} We integrate newly tokenized structures --- e.g., flow, depth, and segments --- by inserting their tokens between $f_{0}$ and $f_{1}$ in the same autoregressive sequence, following simple sequence-design principles. In this layout, the new tokens serve both as conditioning inputs and as prediction targets, decomposing the scene into core structures that steer the subsequent frame. The full training sequence therefore subsumes the “zoo” of predictors (e.g., motion prediction, flow-conditioned RGB, segment prediction) as masked subsets of a single sequence.}
    
    \label{fig:mixing_in_new_tokens}
\end{figure}

\noindent\textbf{Step 2. Mix in the new tokens.} New tokens are integrated by weaving them into the sequence that $\Psi$ models---we take tokens from existing frames $f_0$ and $f_1$ and interleave the extracted intermediates between them, creating sequences like $[f_0, \text{camera}, \text{flow}, \text{depth}, \text{segments}, f_1]$ (as illustrated in Figure~\ref{fig:mixing_in_new_tokens}). This simple rearrangement defines all the predictors mentioned above ($\Psi_1$, $\Psi_2$, $\Psi_3$, ...) as different prompts of the same unified model, rather than requiring separate specialized architectures. The power of this approach lies in how tasks emerge from sequence structure itself: when RGB and flow tokens precede RGB tokens, we create a flow-conditioned frame predictor; when RGB tokens precede flow tokens, we create a motion predictor. The causal mask enables a single sequence to supervise multiple tasks in superposition---in a sequence $[f_0, \text{flow}, f_1]$, the first flow token sees the model as a motion predictor (conditioned on $f_0$), while the first $f_1$ token sees it as a flow-conditioned frame predictor, all within a single forward pass. This flexibility extends to arbitrary combinations: any subset of tokens can condition the prediction of any other subset~\cite{bai2024sequential}, creating a vast space of possible prompts from a fixed set of integrated structures.
\\

\noindent\textbf{Step 3. Continue training.} Because we employ the Warmup-Stable-Decay (WSD) training schedule, we can introduce new token types during the Stable phase and continue training from the last stable checkpoint seamlessly. The model integrates these new sequences into its training diet without catastrophic forgetting, maintaining performance on existing tasks while learning new capabilities. No architectural modifications or loss function changes are necessary---the model simply continues training with these augmented sequences.
\\

\noindent In the graphical‑model view, the original pointer‑indexed RGB (and known controls like camera pose) are the \emph{observables} \(x\), while the newly added token types (flow, depth, segments) are \emph{latents} \(z\). We introduce a \emph{posterior‑as‑data} integration step: obtain approximate posterior samples \(\tilde z \sim \hat P(z\mid x)\) via the counterfactual prompts of §\ref{sec:structure_extraction}, interleave \((x,\tilde z)\) in the same LRAS sequence, and continue maximum‑likelihood training. This upgrades the model from \(P_\theta(x)\) to the joint \(P_\theta(x,z)=P_\theta(x)\,P_\theta(z\mid x)\). Because LRAS treats all pointer‑addressed variables uniformly, the same parameters amortize a family of conditionals by toggling which tokens are observed versus predicted (e.g., \(p_\theta(z\mid x)\), \(p_\theta(x_S\mid x_{\bar S},z_T)\), \(p_\theta(z_T\mid x,z_{\bar T})\)). Latents thus become first‑class tokens in the joint; iterating with additional structures yields \(P_\theta(x,z_1,\ldots,z_k)\) and a correspondingly richer set of controllable prompts.
\\

\noindent In the subsequent sections we show that integration allows us to capture four key benefits:
\begin{itemize}
\item \textit{Better control surfaces (\S\ref{subsec:better_control_surfaces}):} By directly specifying intermediate structures --- e.g., per-pixel motion --- it is possible to specify desired manipulations much more precisely. 

\item \textit{Better extractions (\ref{subsec:better_extractions}):} The native prediction of intermediates as primary outputs combined with the precise control they enable produces higher-quality structures than post hoc extraction. 

\item \textit{Access to further new structures (\S\ref{subsec:further_new_tokens}):} Intermediates themselves become sources for higher-order structures (e.g., motion entropy from flow predictions) that were previously inaccessible without the intermediate predictor. 

\item \textit{Improved base prediction (\S\ref{subsec:upgraded_predictor}):} factoring prediction through intermediates (e.g.,\ first through predicting flow, and then predicting future RGB from past RGB and flow) decomposes hard problems into simpler steps, resolving different sources of uncertainty separately.
\end{itemize}
Together, these benefits create a virtuous cycle of continual improvement. Better predictions enable cleaner structure extraction, which when integrated enables even better predictions and unlocks previously inaccessible structures. Each cycle enriches $\Psi$'s abilities: pixels lead to motion, motion reveals objects, objects enable affordances, building toward complete scene understanding. This recursive bootstrapping mirrors the hierarchical nature of visual understanding---from pixels to motion to affordances---with each integrated structure expanding the space of queryable properties and enabling second-order extractions that would be impossible from RGB alone.
\\

\noindent\textbf{Key Training Details.} We validate our integration mechanism by integrating optical flow into a $\Psi$ model initially trained only on RGB. We first train an HLQ tokenizer on flows extracted from our video dataset, then continue training $\Psi$ on sequences augmented with flow tokens for an additional 0.5T tokens, resuming from the last Stable phase checkpoint of the RGB model training schedule. The model seamlessly learns to both predict and condition on flow without forgetting its RGB capabilities. The following sections explore how this single integration unlocks precise control, improved extractions, and better base predictions---demonstrating the power of the PSI framework.


\subsection{Exposing Richer Control Surfaces}
\label{subsec:better_control_surfaces}

\begin{figure}[h]
    \centering
    \vspace{-10pt}
    \captionsetup{type=figure}
    \captionsetup{labelfont=bf}
    \includegraphics[width=0.9\textwidth]{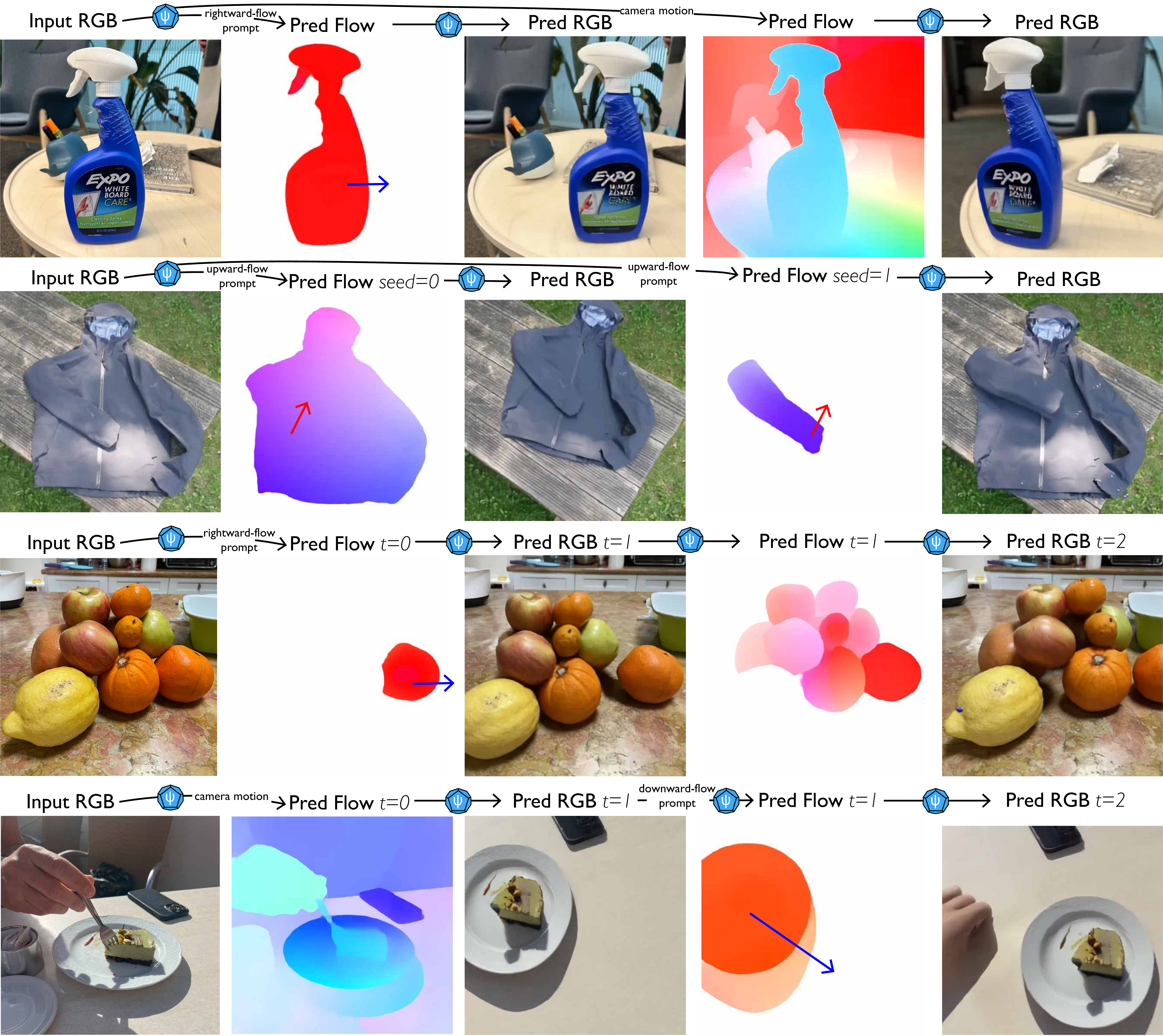}
    \caption{\textbf{Improved Generation Control.} By allowing PSI ($\Psi$) to accept flow tokens as conditioning, we expose a powerful new control surface to guide our generations. By strictly constraining the motion we wish our model to simulate, we significantly constrain the space of plausible outputs.} 
    \label{fig:flow_conditioned_control_qualitative}
\end{figure}

The integration of flow tokens allows us to exercise more precise control over the generations of $\Psi$. For example, we can overcome the fundamental limitations of RGB patch copying from Section~\ref{subsec:inference_pathways}, and instead specify motion through sparse flow vectors on an object as shown in Figure~\ref{fig:flow_conditioned_control_qualitative}. Crucially, these flow specifications can be as sparse as needed---from a single vector to induce translation, to a handful of vectors placed anywhere on the object to specify more complex motions. Let $\mathcal{O}$ be the set of patch indices belonging to the object we wish to move, and $\mathbf{v}_i$ the desired motion vectors. The model propagates these sparse constraints into complete, physically plausible motion fields:
\[
\mathbf{F}_\text{sparse} = \{(\mathbf{p}_i, \mathbf{v}_i) : i \in \mathcal{O}\}, \quad
f_1 \sim \Psi[f_0, \mathbf{F}_\text{sparse}]
\]
For rigid objects, sparse flows anywhere on the object generate coherent motion fields respecting rigidity; for non-rigid materials like fabric (Figure~\ref{fig:flow_conditioned_control_qualitative}), the same sparse flows produce appropriate deformations based on material properties.

Camera control can also be made more precise with flow tokens. When using raw camera pose tokens, the model must infer camera intrinsics and scene scale---leading to a distribution over many plausible interpretations. With flow tokens, we can compute the exact pixel displacements for a given camera motion and directly specify them:
\[
\mathbf{F}_\mathbf{T} = \text{HomographyFlow}(\mathbf{T}, \mathbf{K}, \hat{\mathbf{D}}), \quad
f_1 \sim \Psi[f_0, \mathbf{F}_\mathbf{T}]
\]
where $\mathbf{T}$ is the transformation, $\mathbf{K}$ the known camera intrinsics, and $\hat{\mathbf{D}}$ the estimated depth.

\begin{table}[!h]
  \centering
  \renewcommand{\arraystretch}{1.2}
  \setlength{\tabcolsep}{4pt}
  \begin{tabular}{@{}c c@{}}
    \begin{minipage}{0.48\linewidth}
      \centering
      \textbf{WildRGB-D: Novel View Synthesis}\\[2pt]
      \begin{tabular}{lccc}
        \toprule
        \textbf{Model} & \textbf{PSNR} $\uparrow$ & \textbf{SSIM} $\uparrow$ & \textbf{LPIPS} $\downarrow$\\
        \midrule
        MotionCtrl & 12.39 & 0.293 & 0.404 \\
        ZeroNVS & 16.14 & 0.460 & 0.283 \\
        ViewCrafter & 13.96 & 0.375 & 0.290 \\
        \textbf{PSI$_{rgb}$} & 14.49 & 0.389 & 0.346 \\
        \textbf{PSI} & \textbf{18.27} & \textbf{0.566} & \textbf{0.177} \\
        \bottomrule
      \end{tabular}
    \end{minipage}
    &
    \begin{minipage}{0.48\linewidth}
      \centering
      \textbf{3DEditBench: Object Manipulation}\\[2pt]
      \begin{tabular}{lccc}
        \toprule
        \textbf{Model} & \textbf{PSNR} $\uparrow$ & \textbf{LPIPS} $\downarrow$ & \textbf{EA} $\uparrow$ \\
        \midrule
        DragAnything & 15.13 & 0.443 & 0.517 \\
        Diffusion Handles & 17.82 & 0.344 & 0.619 \\
        LightningDrag & 19.52 & 0.184 & 0.722 \\
        \textbf{PSI} & \textbf{22.73} & \textbf{0.133} & \textbf{0.797} \\
        \bottomrule
      \end{tabular}
    \end{minipage}
  \end{tabular}
  \caption{\textbf{Quantitative results on novel view synthesis and object manipulation.} Left: On WildRGB-D, PSI with flow-based camera control outperforms specialized novel view synthesis models. Right: On 3DEditBench~\cite{spelke_net}, PSI achieves state-of-the-art performance for physically plausible object manipulation, significantly outperforming diffusion-based editing methods.}
  \label{tab:nvs_and_editing}
\end{table}

We evaluate these capabilities on two complementary benchmarks. WildRGB-D tests novel view synthesis on challenging real-world scenes with large viewpoint changes---a task that requires accurate 3D understanding and realistic disocclusion synthesis. 3DEditBench~\cite{spelke_net} evaluates physically plausible object manipulation through complex 3D transformations like rotations and inter-object occlusions, measuring both visual quality (PSNR, LPIPS) and whether the edited object lands in the correct location (Edit Adherence). As shown in Table~\ref{tab:nvs_and_editing}, $\Psi$ achieves state-of-the-art performance on both tasks. For novel view synthesis, our flow-based camera control significantly outperforms specialized models. For object manipulation, PSI surpasses all diffusion-based editing methods by a large margin, achieving 22.73 PSNR and 0.797 Edit Adherence. These results validate that separating motion from appearance through intermediate tokens enables both more precise control and better generation quality. Further analysis of these capabilities appears in~\cite{lee20253d}.


\subsection{Improving Structure Extractions}
\label{subsec:better_extractions}

The integration of new tokens fundamentally enhances $\Psi$'s ability to extract structures in two distinct ways. First, the precise control surfaces discussed in the previous section enable more targeted causal inference prompts. Our original approach to object segmentation (Section~\ref{sec:rgb_segmentation}) relied on copying RGB patches to induce motion---an approach that conflated appearance changes with actual movement. With flow tokens integrated, we can now provide pure motion hypotheticals: applying a single sparse flow vector to an object isolates its motion without introducing appearance artifacts. Second, by making optical flow itself a prediction target, we directly query for motion in its native representation. When the model predicts which pixels will move in response to our prompts, it outputs motion fields that inherently separate movement from lighting changes or shadows. This yields dramatically cleaner segmentations, as we're isolating exactly the property that defines objects---their tendency to move coherently as unified entities.

Figure~\ref{fig:improved_extractions} demonstrates how segmentation can fail in two ways without integration: either the flow predictor must compute flow between RGB frames (introducing errors), or the model moves objects along with their shadows, failing to isolate physical boundaries. Flow-based prompts solve both issues by encoding pure motion---a natural consequence of this is that we can isolate shadows as the difference between RGB and flow hypotheticals, since RGB captures all appearance changes while flow captures only physical motion.

Camera-conditioned flow generation similarly improves depth extraction by outputting predictions in their native motion representation. Since depth manifests as differential motion---closer objects exhibit larger apparent motion than distant ones during camera movement---flow tokens directly encode this parallax signal, yielding sharper and more precise depth boundaries. 

Quantitatively, these improvements are substantial. As shown in Table~\ref{tab:improved_extractions}, flow-based segmentation achieves 0.57 mIoU on SpelkeEntitySeg, significantly outperforming both RGB-based methods and specialized segmentation models. For depth extraction, PSI achieves state-of-the-art self-supervised performance across multiple benchmarks, with particularly strong results on dynamic scenes where motion cues are essential. The key insight is that integrated flow tokens enable direct prediction of motion-based structures without the indirection of RGB generation, yielding both better quality and computational efficiency. This validates the core principle: the right intermediate representation fundamentally changes what the model can extract.

\begin{figure}[H]
    \centering
    \captionsetup{type=figure}
    \captionsetup{labelfont=bf}
    \includegraphics[width=0.85\textwidth]{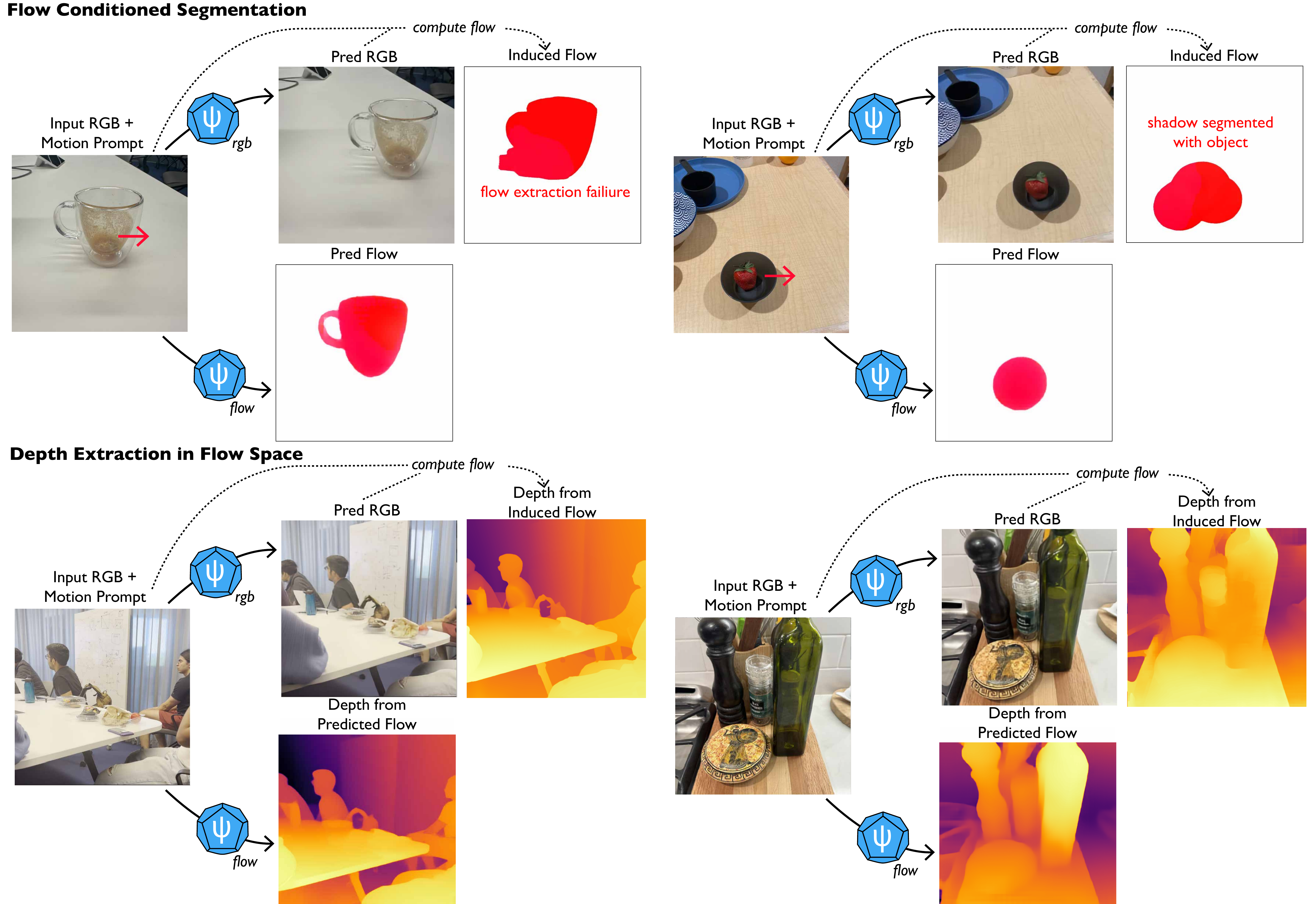}
\end{figure}

\afterpage{
  \captionsetup{type=figure,labelfont=bf}
  \captionof{figure}{\textbf{Improved Extractions Through Direct Intermediate Prediction.} By including intermediate extraction tokens into the $\Psi$ sequence modeling problem, we allow the model to be both conditioned on and predict the intermediates. Here we illustrate how flow can be used as conditioning to apply more precise motion edits to obtain cleaner object segments (top two rows), bypassing off-the-shelf flow extractor errors and ambiguous segments (such as objects). We also illustrate how directly predicting the optical flow field can improve the depth prediction by sidestepping the failure modes of off-the-shelf flow predictors and obtaining sharp, detailed depth estimates.}
  \label{fig:improved_extractions}
}

\begin{table}[h]
  \centering
  \renewcommand{\arraystretch}{1.1}
  \setlength{\tabcolsep}{3pt}
  \resizebox{\linewidth}{!}{%
  \begin{tabular}{@{}l*{4}{c}@{\hspace{10pt}}l*{4}{c}@{}}
    \toprule
    \multicolumn{5}{c}{\textbf{SpelkeBench: Auto-Segmentation}} &
    \multicolumn{5}{c}{\textbf{Monocular Depth Estimation} ($\delta_1$)} \\
    \cmidrule(lr){1-5}\cmidrule(lr){6-10}
    & SAM & DinoV2-G/14 & \textbf{PSI$_{rgb}$} & \textbf{PSI} &
    & MotionCtrl & SC-DepthV2 & IndoorDepth & \textbf{PSI} \\
    \midrule
    \textbf{mIoU} $\uparrow$ & 0.62 & 0.46 & 0.58 & \textbf{0.65} &
    \textbf{NYUD} $\uparrow$ & 0.624 & 0.819 & 0.857 & \textbf{0.873} \\
    \textbf{AR} $\uparrow$ & 0.48 & 0.23 & 0.43 & \textbf{0.54} &
    \textbf{BONN} $\uparrow$ & 0.798 & 0.800 & 0.827 & \textbf{0.889} \\
    \bottomrule
  \end{tabular}}
  \vspace{-5pt}
  \caption{\textbf{Improved structure extraction with integrated flow tokens.} Auto-segmentation on SpelkeBench (left block: mIoU and AR) and self-supervised depth ($\delta_1$) on NYUD and BONN (right block). PSI discovers physically coherent segments that outperform self-supervised methods and achieves state-of-the-art depth, particularly on dynamic scenes where traditional methods fail.}
  \vspace{-10pt}
  \label{tab:improved_extractions}
\end{table}


\subsection{Bootstrapping Further New Tokens}
\label{subsec:further_new_tokens}

The ability to model intermediate structures directly unlocks access to higher-order properties that emerge from the statistics of these intermediates themselves. A concrete example of which, we introduce probability of motion ($P_\text{motion}$)---a property that emerges only after flow becomes part of $\Psi$'s vocabulary. Unlike first-order extractions that reveal what is currently happening (like optical flow showing active motion), $P_\text{motion}$ captures what is likely to happen: the probability that each patch will move in the immediate future, before any motion has occurred. As shown in Figure~\ref{fig:pmotion}, this reveals striking patterns: in static scenes (top row), movable objects like cups and toys show higher motion probability than walls or floors; in kinematic scenes (middle row), objects with implied motion exhibit extreme probability concentrations; for articulated objects (bottom row), the model even captures movement hierarchies---a dog's head and tail show higher values than its torso.
By querying the flow model's beliefs about future motion, we compute a statistical hypothetical---aggregating over all plausible motions that might occur:
\[
P_\text{motion}(\mathbf{p}) = \sum_{\|\mathbf{v}\| > \epsilon} \Psi(\mathbf{F}_\mathbf{p} = \mathbf{v} \mid f_0)
\]
where we sum the probability mass assigned to all non-zero flow vectors $\mathbf{v}$ at each spatial position $\mathbf{p}$. This statistical aggregation captures the model's understanding of scene dynamics before any motion occurs. Unlike a single hypothetical that asks "what if this moved?", we're computing the expectation over all possible futures---revealing which pixels are poised for motion based on learned priors about how similar scenes typically evolve.

\begin{figure}[H]
    \centering
    \includegraphics[width=0.95\textwidth]{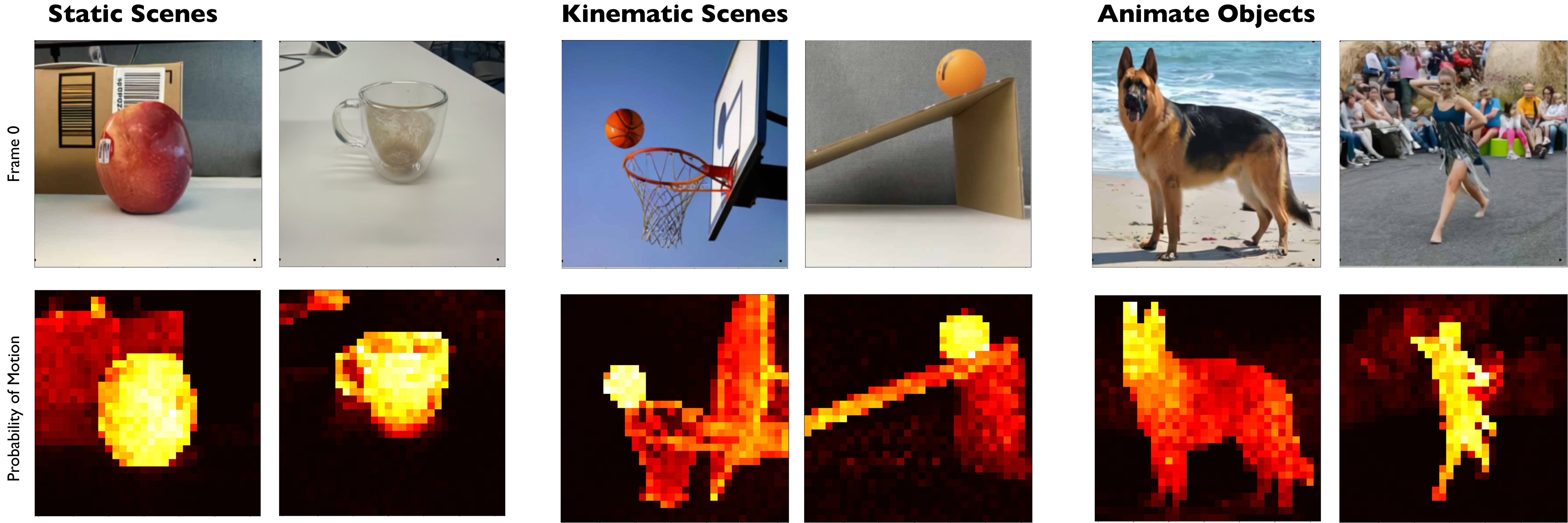}
\end{figure}

\afterpage{
  \captionsetup{type=figure,labelfont=bf}
  \captionof{figure}{\textbf{Probability of Motion.} A parallel prediction is generated using $\Psi_{flow}$ conditioned on the input image. Instead of sampling from the logits, the cumulative probability of all tokens corresponding to any amount of motion is reported. This yields a 2D heatmap of the objects in the image that are likely to move. Top row: static scenes where objects show higher motion probability than backgrounds and surfaces. Middle row: kinematic scenes with obvious implied motion, where active objects have extreme motion probability. Bottom row: animate objects exhibiting high motion probability, with fine-grained structure like the dog's head showing higher values than its body. Note that heatmaps are relative---static scenes have far lower absolute motion probability than dynamic ones.}
  \label{fig:pmotion}
  \vspace{5pt}
}

Beyond binary motion probability, we can extract richer statistics from the flow model's predictions. Expected displacement ($\mathbb{E}[\mathbf{v}]$) captures not just whether motion might occur, but the most likely direction and magnitude at each spatial location:
\[
\mathbb{E}[\mathbf{v}](\mathbf{p}) = \sum_{\mathbf{v}} \mathbf{v} \cdot \Psi(\mathbf{F}_\mathbf{p} = \mathbf{v} \mid f_0)
\]
where we compute the expectation over all possible flow vectors $\mathbf{v}$ weighted by their predicted probabilities. This reveals dominant motion patterns before they occur---a ball at rest shows expected displacement toward the ground due to gravity, a person mid-stride shows forward displacement, and rotating objects exhibit circular displacement fields. Unlike $P_\text{motion}$ which captures motion likelihood, expected displacement encodes the model's beliefs about motion direction and magnitude, providing a vector field of anticipated dynamics that can inform planning and prediction tasks.

The progression from RGB to flow to motion probability illustrates a general principle: each integrated structure enables extraction of properties one level higher in the causal hierarchy. Flow tokens enable extraction of $P_\text{motion}$ and refined object segments; integrated depth could enable extraction of surface normals, material properties, and 3D meshes; segments combined with flow could reveal functional relationships and object permanence. This recursive process creates an open-ended path where each new token type serves as both a useful representation and a stepping stone to richer abstractions. Through this bootstrapping, we can extract increasingly sophisticated scene representations---complete with geometry, dynamics, and affordances---from simple image inputs.


\subsection{Upgrading the Underlying Predictor}
\label{subsec:upgraded_predictor}

One of the most important impacts of structure integration appears in $\Psi$'s original task: RGB frame prediction. By decomposing complex video prediction into structured intermediate steps, we prevent a critical failure mode that plagues direct RGB models: motion collapse. When faced with ambiguous scenarios---a ball at the peak of its trajectory, a person mid-stride, a horse about to jump---models that predict RGB directly often hedge their bets by generating static frames.

Flow tokens force the model to commit to specific motion hypotheses. When predicting RGB directly, the space of all possible future frames---encompassing every combination of motion and appearance change---is so vast that models often assign most probability mass to the safe "no motion" prediction. In contrast, flow space focuses solely on plausible motions, a much more manageable distribution to model. This narrower scope prevents collapse to the static mode: the model can effectively distribute probability across actual motion patterns rather than defaulting to zero. The two-stage process---first predicting flow, then RGB conditioned on flow---transforms an intractable high-dimensional problem into two simpler sequential problems.

This factorization yields concrete improvements. Scenes that contain clear motion cues but produce static outputs in direct prediction models generate appropriate dynamics when motion is explicitly resolved first. Walking humans produce walking motions, galloping horses continue their gait, thrown objects follow ballistic trajectories. Rather than collapsing to static predictions, the model generates diverse samples that each commit to specific motion hypotheses. This captures uncertainty properly---through variation across different samples rather than through averaging motion directions to zero within individual predictions.

Quantitatively, PSI with flow integration outperforms both its RGB-only variant and the COSMOS baseline on VID across multiple datasets (Table~\ref{tab:video_prediction}).  Notably, while COSMOS---trained without conditioning on text prompts---struggles to infer scene dynamics from visual information alone, PSI leverages its flow intermediate to capture motion patterns directly from perceptual data. This demonstrates a fundamental advantage of our approach: by giving the model the vocabulary to reason explicitly about motion, we enable it to extract and utilize dynamic information that remains inaccessible to models limited to pixel-level representations.
 
\begin{figure}[H]
    \centering
    \captionsetup{type=figure}
    \captionsetup{labelfont=bf}
    \includegraphics[width=0.8\textwidth]{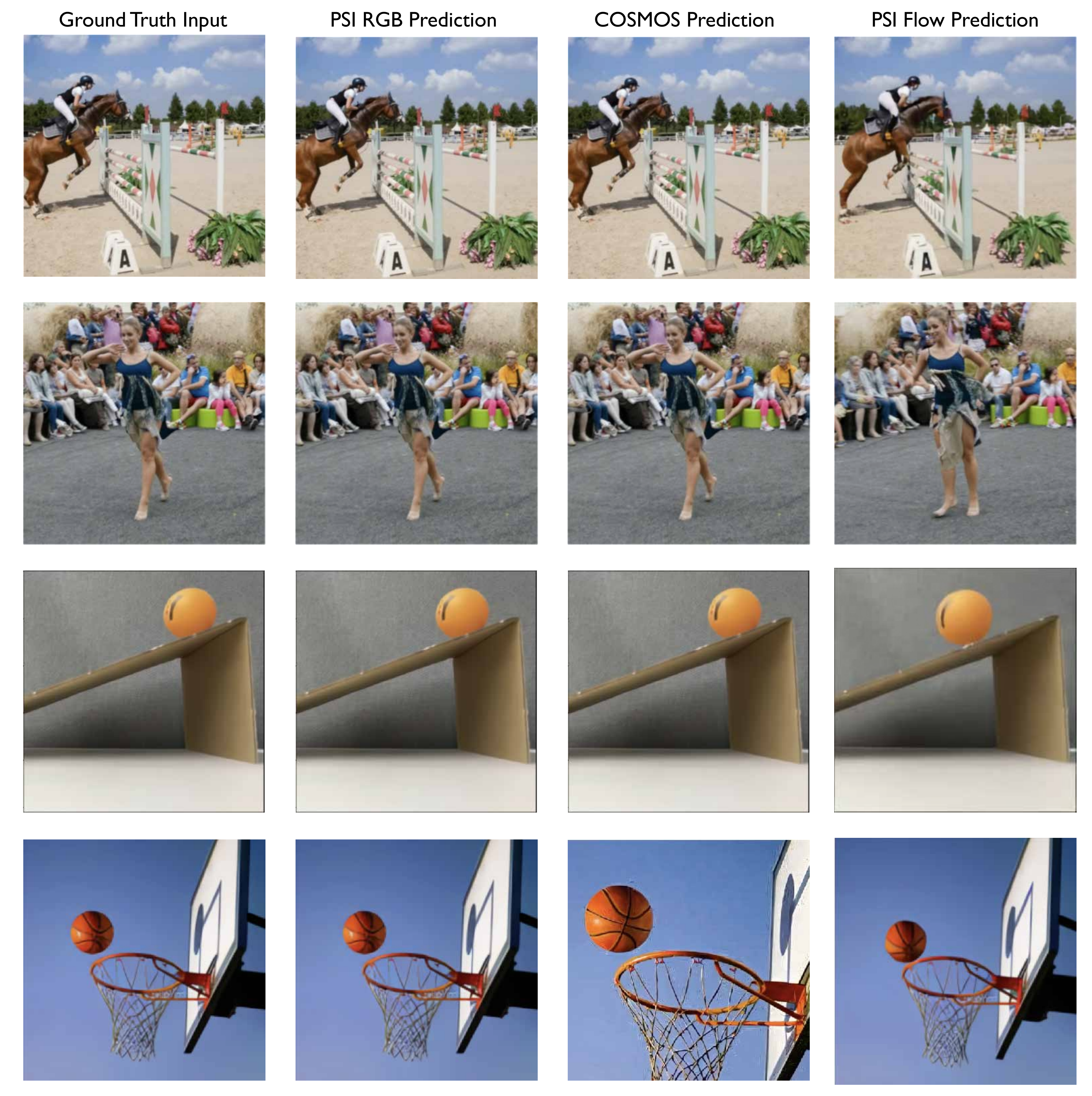}
    \caption{\textbf{Unconditional Prediction with Flow Intermediate.} Single-frame predictions from different models. Columns show: (1) input Frame 0, (2) PSI RGB-only prediction, (3) COSMOS 4B baseline, (4) PSI with flow intermediate, (5) ground truth Frame 1. Both RGB-only models exhibit motion collapse---generating static frames despite obvious motion cues. With flow integration, $\Psi$ first predicts a motion field from Frame 0, then conditions RGB generation on this flow, successfully capturing dynamic scenes that stump direct prediction approaches.}
    \label{fig:unconditional}
\end{figure}

\begin{table}[H]
  \centering
  \renewcommand{\arraystretch}{1.2}
  \setlength{\tabcolsep}{8pt}
  \begin{tabular}{lccc}
    \toprule
    \textbf{Model} & \textbf{DAVIS VID} $\downarrow$ & \textbf{YouTube VID} $\downarrow$ & \\
    \midrule
    COSMOS AR 4B & 210 & 340  \\
    PSI 7B (RGB Only) & 223 & 307  \\
    \textbf{PSI 7B (With Flow)} & \textbf{198} & \textbf{173} \\
    \bottomrule
  \end{tabular}
  \caption{\textbf{Video prediction quality with and without flow integration.} We evaluate single-frame video prediction across three diverse datasets: DAVIS~\cite{Perazzi2016} and YouTube---real-world videos containing interesting motion patterns. For each benchmark, models are conditioned on a single frame and asked to predict the frame 500 ms in the future.}
  \label{tab:video_prediction}
\end{table}

%% file: sections/applications.tex
\section{Example Applications}
\label{sec:applications}

Having described the PSI cycle in detail across the previous three sections (\S\ref{sec:prediction}--\ref{sec:integration}), we now showcase practical applications of the model to existing problems in computer vision and physical reasoning. Our goal is illustrative rather than exhaustive: to demonstrate how a single prompting interface enables factual rollouts, counterfactual analysis, and physically consistent manipulation across diverse tasks.

\subsection{Physical Video Editing with PSI}
\label{subsec:phys_video_edit}

\noindent\textbf{PSI as a physical prediction engine for video editing.}
By conditioning on a small set of context frames, $\Psi$ rolls out futures that respect contact, momentum transfer, and gravity; because the context is editable, localized pixel‑level interventions (e.g., translating an object or toggling contact) act as low‑level controls that deterministically recompute downstream effects---enabling factual, hypothetical, and counterfactual generation. In our bowling example, conditioning on two ground‑truth frames yields a factual prediction that closely matches the true next frame---the ball strikes the pin and the pin topples---demonstrating that $\Psi$ captures contact physics, while the same mechanism supports counterfactual edits that redirect the ball and prevent the pins from falling.

\begin{center}
\includegraphics[width=0.9\textwidth]{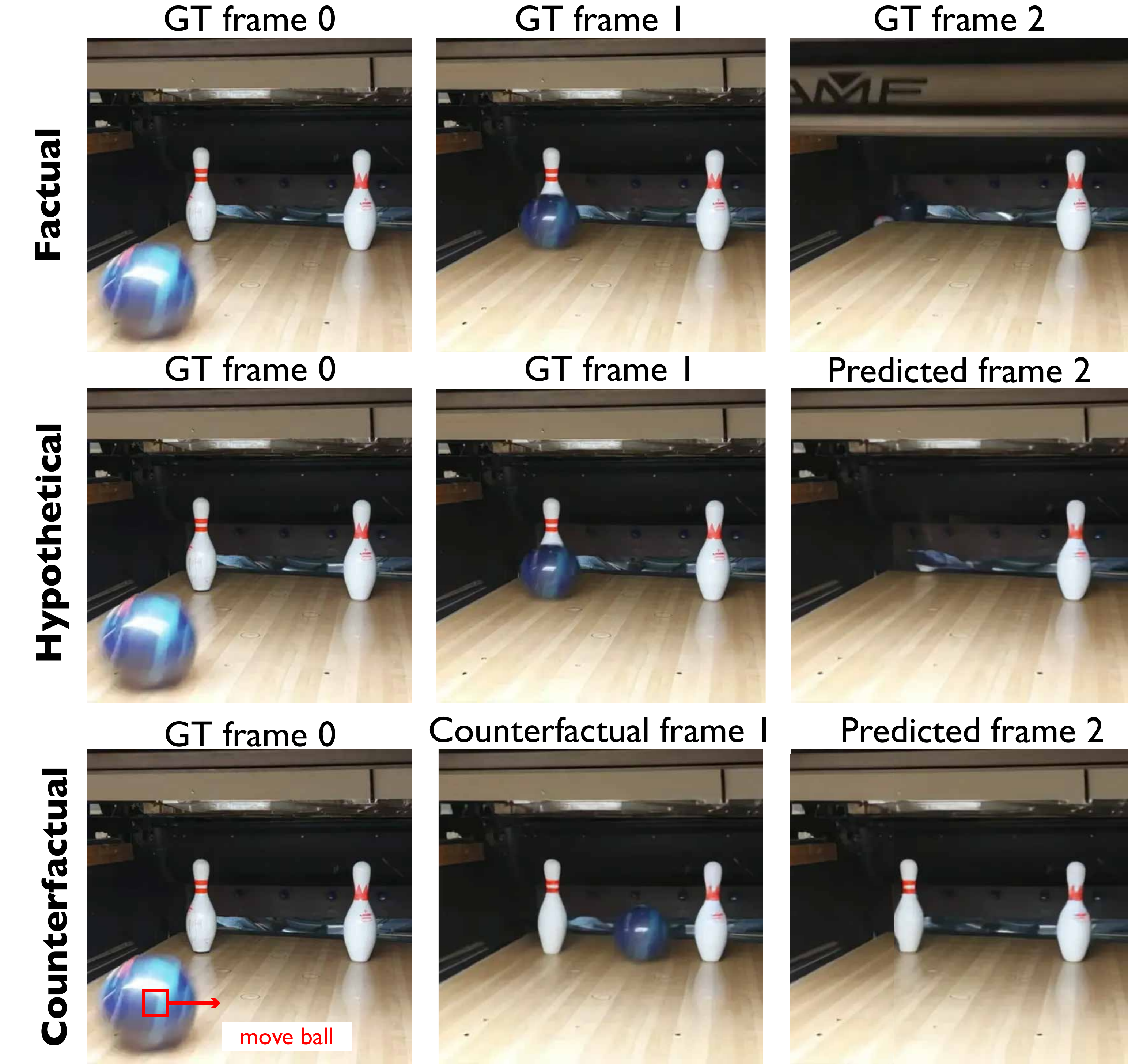}
\end{center}

\medskip
\noindent\textbf{Intervene, then recompute the future.}
With a localized edit (we translate the ball mid‑trajectory in the last context frame), PSI produces a \emph{counterfactual} where the ball misses, so no force is transferred and the pins remain upright. In causal notation, this translation is a $\mathrm{do}$‑operation on the local (patch) variables controlling the ball’s position at the edited time step, and the model propagates the consequences through the learned dependencies. This demonstrates controllable, physically consistent video manipulation: small edits to inputs yield coherent, downstream changes to outcomes.

\subsection{Visual Jenga}
\label{subsec:visual_jenga}

The Visual Jenga task~\cite{bhattad2025visualjenga} requires identifying which object in a stacked scene can be safely removed without causing others to fall. We adopt the object segmentation approach from prior work but replace inpainting‑based stability assessment with a physics‑grounded solution using our flow‑integrated model.

\paragraph{Bidirectional dependency testing.}
For each pair of objects $(A,B)$ in the scene, we perform two complementary tests: (1) prompt object $A$ with a small motion vector and measure the probability that object $B$ moves in response, and (2) reverse the test by prompting $B$ and measuring $A$'s response probability. This yields an asymmetric dependency score $D(A \rightarrow B) \neq D(B \rightarrow A)$, capturing which object's motion more strongly influences the other.

\paragraph{Amodal completion through flow.}
Rather than using inpainting to remove objects---which can introduce unrealistic replacements or fail to respect physics---we leverage large‑magnitude optical flow to achieve true amodal completion~\cite{lee20253d}. By applying flow vectors sufficient to translate an object entirely out of frame, we force the model to reveal occluded scene content while maintaining physical consistency. This approach ensures complete object removal without the hallucination artifacts common in inpainting methods.

\medskip
\noindent\includegraphics[width=\textwidth]{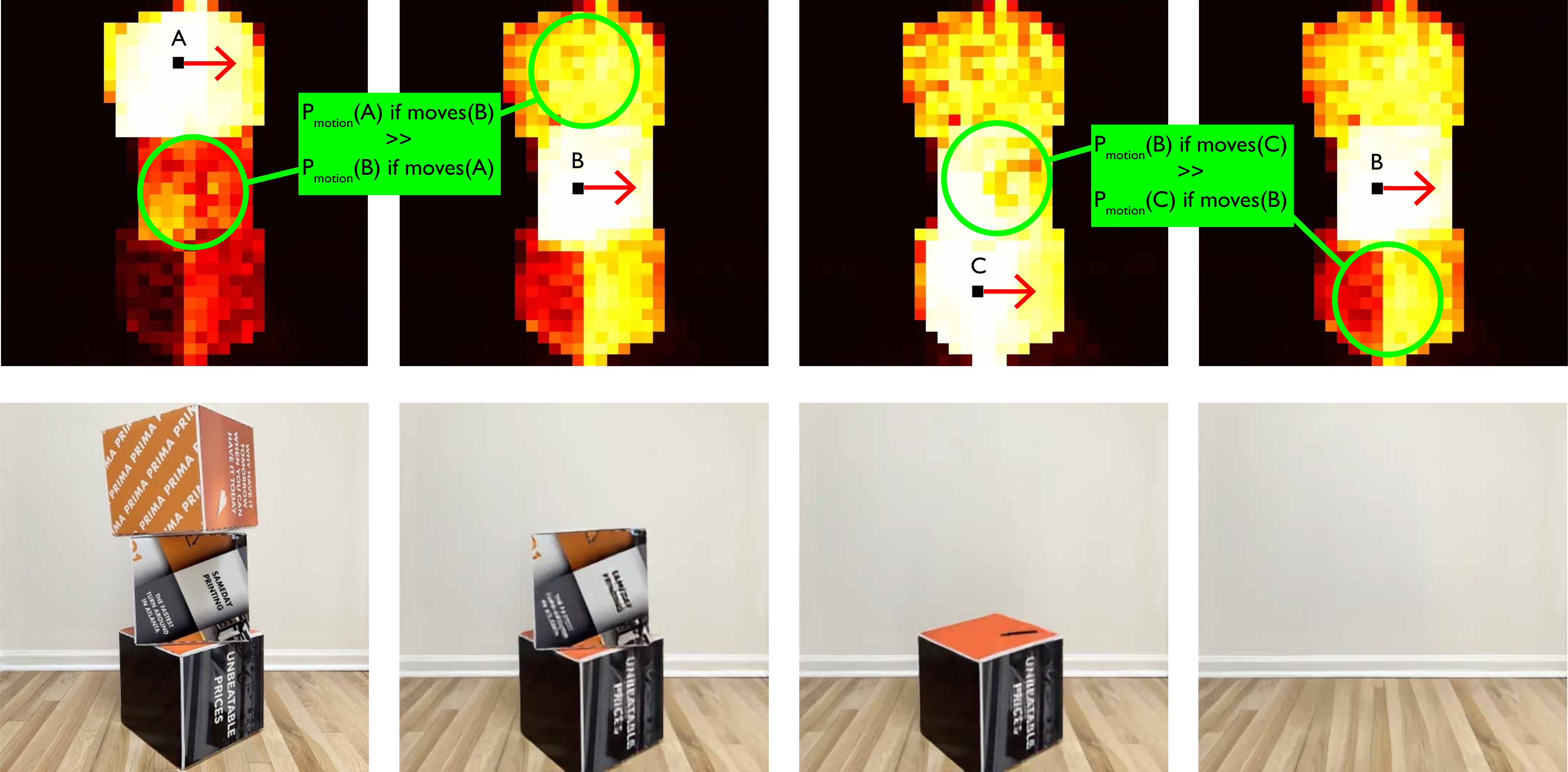}

Combining bidirectional dependency testing with flow‑based amodal completion yields a robust solution to Visual Jenga: we identify the object with minimal outgoing dependencies (least likely to cause others to move when removed), then verify stability by applying amodal completion and checking that remaining objects maintain their positions. This physics‑aware approach significantly outperforms inpainting baselines by respecting the causal structure of object interactions.

\subsection{Robotic Motion Map}
\label{subsec:robotic_motion_map}

Robotic manipulation tasks require understanding not just what objects are present, but how they might move when interacted with. We demonstrate how PSI's probability of motion ($P_\text{motion}$) predictions enable robots to anticipate dynamics before any motion occurs, simply by analyzing single‑frame images.

\paragraph{Anticipating motion from static images.}
A key insight from our flow integration is that PSI learns to predict which objects are likely to move \emph{before} any motion has occurred. By querying $\Psi_1$ with a single input frame $f_0$, we extract probability‑of‑motion maps that encode $P(\text{motion}\mid f_0)$---the likelihood that each patch will move given only the current static scene configuration. This capability emerges from the model's understanding of physical dynamics learned from video data.

\paragraph{Motion probability for robotic planning.}
For a robot manipulating objects on a tabletop, these motion‑probability predictions from single frames provide crucial planning information. Objects with high motion probability are identified as manipulable, while low‑probability regions indicate stable surfaces or immovable obstacles. The robot can query these probability fields to identify which objects are most likely to respond to manipulation, all from a single static observation. This anticipatory understanding---predicting what will move before applying any force---enables more intelligent grasp selection and manipulation strategies.

\medskip
\noindent\includegraphics[width=\textwidth]{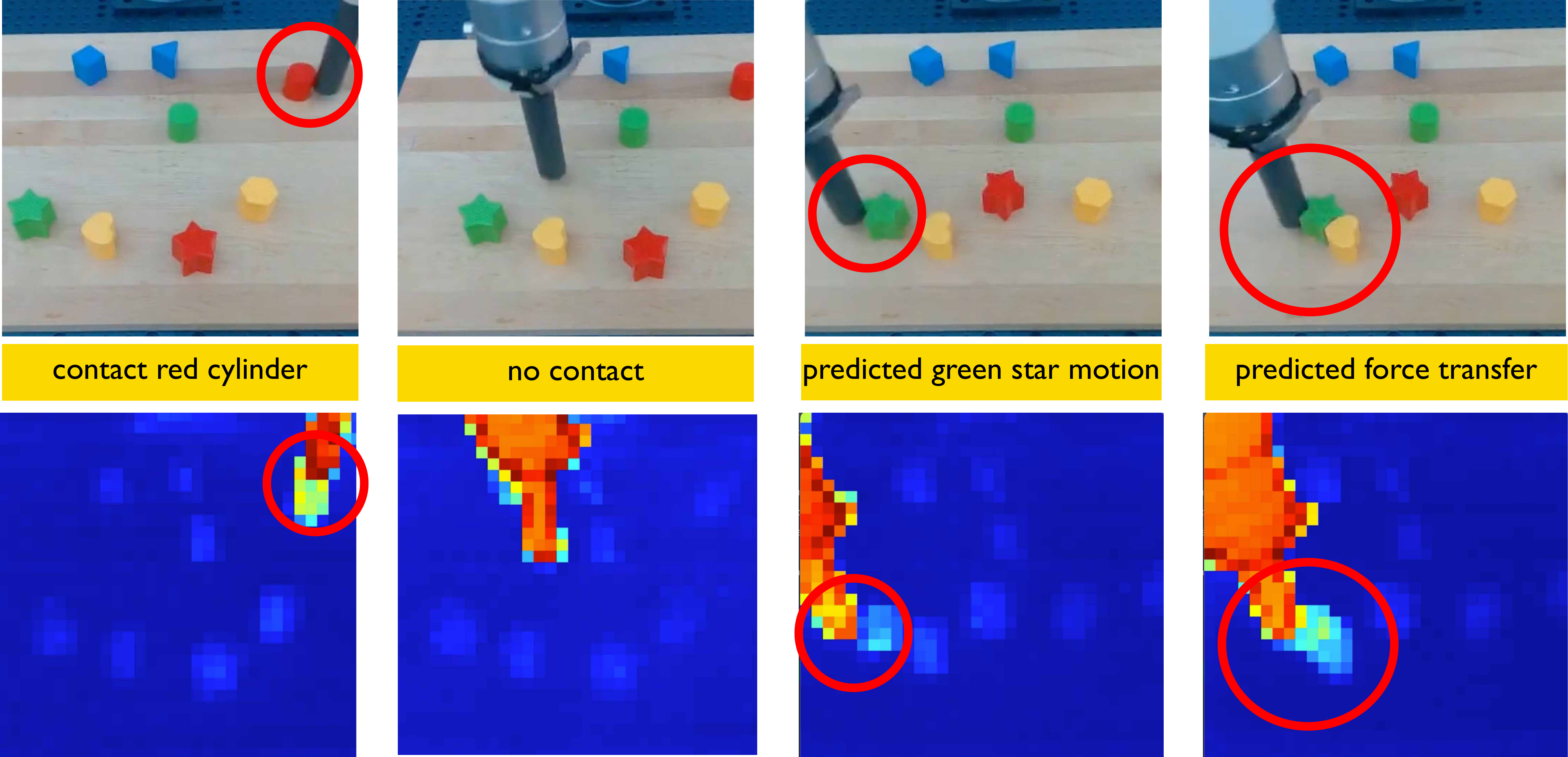}

This single‑frame motion prediction requires no interaction or motion observation to identify manipulable objects, generalizes to novel object configurations without task‑specific training, and provides uncertainty estimates critical for safe manipulation. By anticipating motion from static observations, robots can plan more effectively before executing any actions.

%% file: sections/discussion.tex
\section{Discussion}
\label{sec:discussion}

The PSI framework introduces several ideas.  In summary:
\begin{enumerate}
    \item By creating pointer-indexed local token sequences, and training an autoregressive model on these sequences in random order, we can create extremely scalable and highly controllable generative world models for complex real-world data, such as natural videos.  These models act as full-fledged learned probabilistic graphical models of their data domains.  

    \item Useful quantities (also known as intermediate structures) are zero-shot causal inferences in the base predictor model, implemented via simple analysis programs that compare hypothetical (often counterfactual) predictions with ground truth observations.

    \item The new ``higher-order'' token types emitted from step 2 can be integrated back into training data sequences using an utterly simple and generic token intermixing procedure, augmenting the base predictor with richer control surfaces and higher-fidelity prediction capacities.
\end{enumerate}
Together, these ideas constitute a comprehensive framework for building powerful models of complex natural data domains.  We illustrate these ideas in the domain of natural video. 

The next sections link PSI to a variety of approaches in the literature, aiming to sketch a broader perspective on how these ideas interconnect. 

\subsection{Beyond Representation Learning: From Probes to Prompting to Integration}
\label{subsec:probe_to_prompt}
Representation learning is the idea of pretraining a network for one proxy objective, and then using the features arising from that optimization as a basis for supporting the learning of many other tasks \cite{LeCun2015Deep}. Representation learning works on the principle that, if the proxy task is sufficiently broad, and the pretraining dataset sufficiently varied, the features needed to solve it will likely be useful for many other purposes. Initially, the proxy training objective was a supervised task (e.g.\ ImageNet categorization~\cite{deng2009imagenet}), but the prohibitive cost and task-specific bias of large annotated corpora motivated a shift toward self-supervised objectives, including contrastive embeddings~\cite{He2020Momentum,Chen2020Simple}, masked predictors~\cite{Devlin2019BERT, He2022Masked, tong2022videomae}, and autoencoders~\cite{Higgins2016BetaVAE}, which seek in various ways to align the learned model's latent variables with the environment’s generative factors. Representation learning has proven very powerful and flexible, and applicable to pretrained models of many kinds, including modern generative AI~\cite{Ramesh2022DALLE2}.

The major limitation of representation learning is that it requires \emph{probing} to answer any specific question -- that is, the supervised training of a probe (often implemented as a linear or shallow MLP) on top of the pretrained representation. While this is less laborious than supervised pretraining of the base model, it is nonetheless dramatically less efficient and flexible than \emph{prompting}, the paradigm associated with modern LLMs.

The LLM prompting paradigm has two key properties that distinguish it from representation learning: (a) one single model is sufficiently widely trained that it can itself answer most or all questions of interest, and (b) there is a generic ``format'' for posing all such questions directly to the model as zero-shot conditioners at inference time -- as with natural language prompts. The shift from probing learned representations to zero-shot prompting exhibited in the transition from GPT-2 to GPT-3 is one of the singular achievements of human technology~\cite{Brown2020GPT3}. The flexibility of language prompts to formulate not just simple tasks (such as grammar checks~\cite{Stahlberg2019GEC}) but actually to contain cognitive strategies~\cite{Wei2022CoT} and even define entire domains of intellectual endeavor, such as complex mathematics~\cite{Polu2020GPTf} or social agency~\cite{Park2023GenerativeAgents}, has remained a continual source of surprise and delight. Recent boundary-pushing work has taken this idea full cycle, using the outputs of inference-time constructs as a basis for creating synthetic training data for further base prediction-model improvements~\cite{Wang2022SelfInstruct}.

While prompting is a tremendously attractive concept, generalizing the prompting paradigm to the many domains where language-based descriptions are sparse or absent -- such as video understanding and robotics -- has proven challenging, both because it has remained difficult to extract quantities of interest in a zero-shot fashion from pretrained models, and because even conceptualizing how to formulate a ``language'' of prompts for such domains is unclear.

One major utility of the current work (but really starting with CWM~\cite{Bear2023CWM}) is a route for beginning to shift non-linguistic domains toward a prompting strategy. The core insight was that a simple generic type of construction -- counterfactual programs -- could zero-shot extract most notions of interest, as long as the base predictor was properly formulated. The first two parts of PSI (probabilistic prediction and the structure extraction) can be thought of in analogy to the LLM case as (i) providing a generic approach to base predictors that, aside from their utility as highly controllable generative models, support (ii) zero-shot ``prompts'' that are highly effective. The third part of PSI, (iii) integration, exhibits a mechanism for using the zero-shot inference-time outputs as conditioners and targets to improve the base model, in analogy (but perhaps more generic) to the way synthetic data has recently begun bootstrapping LLM pretraining.

\textbf{So why not just use text conditioning?}
The analogies we make to LLMs beg the obvious question: if language is so good, why not use it in your world models? Of course, that probably is a good idea, one that many others have followed up~\cite{Ahn2022SayCan,Brohan2023RT2,Driess2023PaLM-E}, and which in the future will almost certainly continue to be of great import.
Here, however, we have taken a different route. At one level this is because, as has become increasingly realized in recent literature from multiple domains~\cite{Ebert2018VisualForesight,Nair2022R3M}, solving many fine-grained and subtle problems of embodiment requires understanding that is not easily text-supervised given the nature of large-scale data available in the real world. Somehow, we will have to come to grips with fine-grained discovery and control of non-text data structures.
However, there is a deeper reason why we have at least temporarily chosen to avoid text. While text conditioning might act as a strong signal for future prediction, it also gives the model a shortcut that is not always present in the real world. By forcing our model to infer scene transformations and motions purely from the input frame, we confront it with the same world-modeling prediction problem that intelligent life effectively solves, forcing it to learn a strong encoder function. It is possible that these non-textual solutions underlie the way cognition is learned in intelligent but pre-linguistic agents such as non-human primates and human infants~\cite{Krupenye2016ApesFalseBelief, Spelke2007CoreKnowledge}.
Ultimately, we do envision integrating language conditioning into PSI to provide an additional control surface--but only after first establishing sufficient visual intermediates (flow, depth, segments, meshes, and beyond) that can capture the full richness of scene dynamics without relying on the coarse abstractions that text necessarily imposes.

\subsection{Related Work}

\textbf{Generative Models \& Diffusion.} Likelihood-based variational autoencoders \cite{Kingma2014VAE} and adversarially trained GANs \cite{Goodfellow2014GAN} first demonstrated scalable generative modeling, but encountered quality–diversity trade-offs for high-resolution imagery.  
Denoising diffusion models address these limitations through iterative refinement \cite{Ho2020Diffusion}; latent diffusion substantially reduces computational cost while preserving fidelity \cite{Rombach2022LDM}.  
In discrete domains, autoregressive Transformers have shown very strong progress: ImageGPT \cite{Chen2020ImageGPT} first extended next-token factorization to pixels, and contemporary large-scale language models \cite{Brown2020GPT3} achieve strong likelihoods and emergent structure.  

\begin{figure}[h]
    \centering
    \captionsetup{type=figure}
    \captionsetup{labelfont=bf}
    \includegraphics[width=0.98\textwidth]{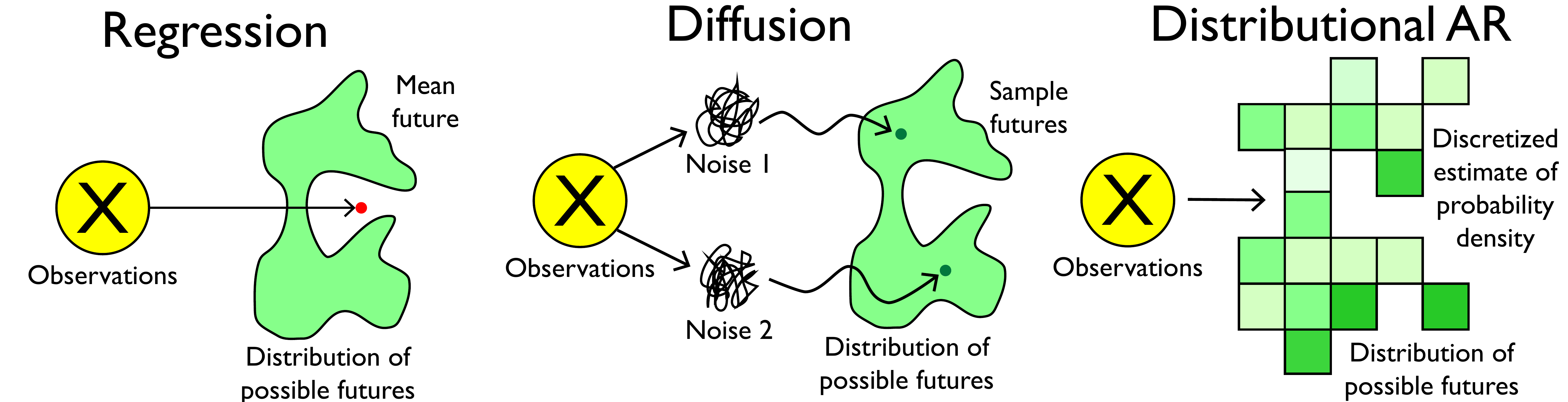}
    \caption{\textbf{Regression, Diffusion, and Distributional Prediction.}
    \emph{Regression models} output the conditional mean of the future, which is often off-manifold -- even though it minimizes distance to most possible futures.
    \emph{Diffusion models} learn continuous paths that transform noise into realistic samples on the future-state manifold, but they do not model the entire probability density explicitly.
    \emph{Distributional autoregressive models} (such as LLMs) output a full probability distribution over a discretized representation of the future, enabling explicit uncertainty estimates.}
    \label{fig:distributional}
\end{figure}
PSI's distributional LRAS-based predictor represents a choice to go all-in on a form of autoregression for generative modeling.  By formulating the prediction problem fully distributionally (Fig.~\ref{fig:distributional}), unlike diffusion models which model one sample at a time, PSI's predictor makes statistical inferences that aggregate information across many potential future trajectories, such as entropy and motion maps, easily accessible.  And despite their successes, both diffusion and existing autoregressive families typically condition generation on global embeddings or prompts. Localized conditioning has received comparatively limited attention but constitutes a focus of PSI.  This choice has the key advantage of enabling precise spatial handles, which is useful both for precision of control at generation time, and for formulating effective hypotheticals for structure extraction.   

\textbf{Random Order Autoregressive Modeling.} Several recent and concurrent works have explored random-order autoregressive modeling, though none combine the three properties that make LRAS uniquely suited for controllable visual generation: random traversal of 2D data, strict token locality, and pointer-based addressing. Sigma-GPT~\cite{ning2024sigmargpt} demonstrates that language models trained with random token permutations achieve comparable perplexity to fixed-order models while enabling flexible generation patterns. However, this work remains confined to 1D text sequences where the notion of spatial locality is less relevant. RandAR~\cite{gao2024randar} brings random ordering to visual domains, training image models with arbitrary raster-scan patterns. Yet RandAR employs VQ-GAN tokenization, where each token encodes global information from a large receptive field—making local edits impossible without affecting distant image regions. XL-VAE~\cite{han2024xlvae} explores random orderings for image generation but focuses on hierarchical coarse-to-fine generation rather than maintaining strict spatial locality. The critical distinction of LRAS is that its combination of strict locality (via HLQ) and pointer-based random access enables the construction of a locally conditioned, queryable probabilistic graphical model.

\textbf{From ``Traditional'' World Models to Data-Driven ``World Models''.} A world model forecasts the consequences of actions.  This concept has a long tradition within model-based reinforcement learning~\cite{Sutton1990Dyna,Deisenroth2011PILCO} and model-predictive control~\cite{Garcia1989MPC}; Ha and Schmidhuber’s \emph{World Models} architecture \cite{Ha2018World}—a VAE coupled with a recurrent dynamics model—provided a simple and understandable toy example of this principle with modern neural networks. PlaNet \cite{Hafner2019PlaNet}, Dreamer \cite{hafner2020dreamer}, and Director~\cite{Hafner2022Director} subsequently improved long-horizon planning by leveraging latent imagination in more sophisticated visual contexts. Recent lines of work have extended these ideas in a variety of directions~\cite{Kaiser2019SimPLe,Schrittwieser2020MuZero,Wu2023DayDreamer, Kotar_2023_ICCV}.  

It might at first seem odd that we've discussed PSI as a framework for world modeling---after all, the inputs to $\Psi$ are just data, so where are the actions whose consequences are to be forecasted?  In a sense PSI (and the CWM concept that came before it) is a ``poor man's world model'', where expensive-to-obtain true action data is often proxied by cheap data patches that encode approximations to simple actions.  That this proxy is effective is evidenced by the way that flow ``tracers'' and segment ``motions'' work well. The inferential form of $\Psi$ formally treats data patches in precisely the same way that true action data would be, and reaps the reward of doing so because, as it turns out, training on raw data enables the underlying model to learn enough about the way the world works that it can competently perform hypotheticals (that is, enact $\texttt{do}$ operators) without being hopelessly out-of-distribution.  

The kind of ``data-driven'' world model represented by PSI has two advantages over traditional world modeling.  First, it can learn most of what it needs to know about the world without the need for explicit action data, dramatically increasing its efficiency as a practical technological approach.  Second, PSI's insistence on describing all predictions through the lens of potential ``actions'' forces a reckoning with the fact that no one observer is the only true agent in the environment; real things in the world outside the ``agent'' have causal status, and it is needlessly limiting to pretend otherwise.  And of course, when real action data is available (as in the case of camera-conditioning discussed in \S\ref{subsec:inference_pathways}), $\Psi$ can seamlessly integrate it. 

\textbf{PSI vs Counterfactual World Modeling.} Bear \textit{et al.} introduce \emph{Counterfactual World Modeling} (CWM) as a vision foundation‐model framework that unifies a broad spectrum of visual tasks without recourse to task‐specific supervision \cite{Bear2023CWM}. CWM rests on two interacting ingredients.  First, a \emph{structured masking} objective—implemented as \emph{temporally-factored masking} in video—presents the predictor with an entire reference frame while masking almost all patches of the subsequent frame.  This asymmetry forces the network to concentrate information about scene dynamics into a small set of visible tokens, effectively disentangling appearance from motion, and giving a strong interface for local control.  Second, \emph{counterfactual prompting} computes diverse mid-level representations by comparing the model’s predictions on the actual input with its predictions on minimally perturbed (``counterfactual") versions of that input.  Without any fine-tuning, a single pretrained CWM yields zero-shot estimates of keypoints, optical flow, occlusion boundaries, relative depth, and object segments.  Subsequent work demonstrates that the same framework can be leveraged for physical reasoning. Venkatesh \textit{et al.}~\cite{venkatesh_counterfactual_2023} show that CWM’s counterfactual queries extract structures—such as motion groups and contact events—that support state-of-the-art performance on the Physion dynamics benchmark, again without annotated training data.  Stojanov \textit{et al.}~\cite{stojanov2025self} show how CWM-style counterfactuals can be optimized to provide extremely effective optical flow extraction. 

PSI solves two problems with CWM. First, CWM's underlying predictor was a regression-based masked autoencoder, leading to the ``blurriness'' associated with severe mode collapse, a limitation that affected many downstream uses of CWM.  PSI rectifies this by replacing the masked autoencoder with a fully distributional predictor capable of true ``uncertainty management''. Second, CWM provided no generic mechanism of integration --- that is, using the outputs of structured extractions to improve the base world model (though the Opt-CWM framework in Stojanov \textit{et al.}~\cite{stojanov2025self} shows one approach to this).  PSI's reformulation in terms of sequence prediction, and the introduction of the sequence mixing concept for integration, rectifies this problem, enabling a fully virtuous cycle. 

The cost of PSI over CWM is model size.  PSI models have to solve a much harder problem than CWM --- predicting a whole distribution rather than the mean tendency of that distribution --- and achieving models that recover CWM functionality  within the PSI framework (while solving many other problems along the way) requires substantially more parameters than in CWM.  Thankfully, the amount of data for training does not need to scale significantly --- bigger PSI models simply get more out of the data than smaller (or even bigger) CWM models can. 

\textbf{Learned Probabilistic and Bayesian Models.}
There have been a number of works seeking to learn PGMs from images and video data~\cite{frey2013learning, beal2003graphical, malgireddy2012temporal}.  More recently, the probabilistic programming community has taken steps to integrate formal Bayesian perspectives on structure representation with learnable models, and applied these ideas to some computer vision problems~\cite{becker2024probabilistic, gothoskar20213dp3, cusumano2019gen}. Similar ideas motivate the addition of program-like structure to language models~\cite{wong2023word,wong2025modeling}.  Better understanding the connection between these works and PSI is also a subject for future investigation. 

\subsection{Predictions, Interventions, Hypotheticals \& Counterfactuals}
Pearl's ladder of causation~\cite{Pearl2009Causality} provides a framework in which to think about causal reasoning.  In that framework, there is a clear distinction between predictions and interventions --- that is, a difference between $p(y | X=x)$ and $p(y | \texttt{do}(x))$ --- relying on the well-definedness of the \texttt{do} operator for modifying (cutting links in) the underlying probabilistic graph of the system.  Physically, applying the \texttt{do} operator corresponds to an exogenous process that changes the system's structural equation model (SEM) in a minimally-invasive fashion. 

In the PSI setting, we are working with learned approximations to a very large PGM, where minimal modifications to the structural equations of the system are less well-defined and exogenous processes are inadmissible. We thus use a semantics of causation that is somewhat different from the classical Pearlian ladder of causation -- following, in part, the suggestions of Tim Maudlin~\cite{maudlin2007metaphysics}. In our setting, \emph{predictions} are the distributional outputs of $\Psi$ when conditioned on ground truth observations, as well as any parallel-mode samples from these predictions; \emph{intervention} is the act of obtaining predictions from $\Psi$ on non ground truth (synthesized) conditioning; and \emph{hypotheticals} are the sequential-model samples from such predictions.  We do not make a bright-line distinction between $p(y | X=x)$ and $p(y | \texttt{do}(x))$, but instead allow for a spectrum of intervention types expressed through different (non ground truth) input conditioning.  The process of integration (\S\ref{sec:integration}), in which improved control surfaces are discovered, can be thought of as a way to ease the construction of ``mini-surgeries'' to the causal graph, converging toward the usual semantics of causal inference.  A similar thought can be applied to methods of counterfactual optimization that achieve improved minimal interventions via a bootstrapping principle~\cite{stojanov2025self}.

Though PSI does not admit exogenous processes (i.e. acts-of-god interventions outside the system dynamics), there is an equivalent in the PSI framework.  As with any learned model, $\Psi$ can be evaluated on unobserved conditioning inputs (that is, interventions in our sense) that are outside the training distribution of the model.  Whether and where $\Psi$ generalizes is an empirical question in each setting.  Failure is detectable and the model can be refined, though not by changing the SEM explicitly --- but rather by adding training data with a few of the right kinds of interventions observed. The hope is that the learned model will often generalize so that correct predictions can be achieved without having to have explicit examples of literally every version of intervention seen in some training observation. 

One might wonder if causal inferences in this new semantics are as certain as they are under the usual ladder semantics. Indeed they are not -- since they rely on building up ``interventions'' whose mechanisms are as certain as possible, but never quite have the absolute certainty that comes with being able to completely cut all the incoming connections to a given node. Even in the most certain case of causal analysis illustrated in this work -- the tracer analysis yielding optical flow -- it is always possible that the tracer will be interpreted by the model as ``schmutz on the camera lens'' and not flow along with the object it is ``on''. Failure to measure the causal flow in this case will be confounded with not flowing at all. (Though this problem is not actually very significant in practice, as shown by decent results achieved with the method presented in \S\ref{subsec:kl_tracing}, it can even be ameliorated in a principled fashion~\cite{stojanov2025self}.) However -- and in this regard we stand with Nancy Cartwright~\cite{cartwright2007causal} -- no \emph{real-world} causal analysis can ever really be perfectly certain. Any physical manifestation of an intervention can only be implemented by systems relying on built-up empirical understandings resembling a learned model. For example, when a physician administers a treatment, even to know that she has picked up a syringe and successfully depressed the plunger to deliver the intramuscular injection  --- and not failed to do so in any number of byzantine ways -- requires her sensorimotor system to have been trained (a key part of development) on the interpretation of visual, auditory, and haptic data, and their correlations with efferent copies of her own muscle discharges. There is causal uncertainty in this system of the same kind, and perhaps not lesser in magnitude, than that in PSI.

It is worth emphasizing the difference between hypotheticals and \emph{counterfactuals}, since PSI does maintain the usual semantics of this distinction --- and because the difference matters practically.  Counterfactuals are a specific subset of causal inferences -- those in which sufficient information can be provided \emph{factually} so that the uncertainty about the effect of any hypotheticals are dramatically reduced, and perhaps eliminated entirely~\cite{gerstenberg2024counterfactual}.  This reduction in uncertainty is how counterfactuals, when they can really be constructed, provide the ability to identify individual-level causal effects~\cite{pearl20103}.  Counterfactuals thus enable stronger causal inferences as compared to ``mere'' hypotheticals~\cite{pearl20103}.  

Not all useful causal inference prompts to PSI are actually counterfactual -- but some are.  In the case of flow extraction (\S\ref{subsec:kl_tracing}), the tracer hypothetical is truly counterfactual.  This is because a decent amount of information about the second frame $f_1$ is provided to $\Psi$ --- in fact, enough information to reduce the uncertainty about object motion to zero, since it is only if the model is told what the actual motions of objects are from $f_0$ to $f_1$ that it can correctly predict where the hypothetical tracer should go.\footnote{Of course, if the provided $f_1$ patch happens to land where the tracer in $f_0$ would go, the hypothetical is quashed; these aberrant situations can be identified and removed by averaging over random choices of $f_1$ information provided; but this is also why we use \emph{just enough} factual information to reduce object motion uncertainty to zero, but no more, so that as little quashing as possible happens.  But the very possibility of this quashing is a sure sign that this is a \emph{true} counterfactual.} For this reason, we are able to strongly identify the individual causal relationship between the $f_1$ point and its antecedent in $f_0$. 
Movable-object segmentation extraction from single frames is, by contrast, not actually counterfactual, because it merely involves comparisons between multiple hypotheticals, and the uncertainty in each individual prediction remains high throughout.  

This contrast is quite instructive to analyze.  The causal inference in flow is stronger than the causal inference in segments: indeed, movable-segment judgments from static images are subject to irreducible ambiguity, since, after all, an object, like a laptop on a table, that looks like it should be independently movable most of the time, can in any one instance actually be immovable (e.g. because someone surreptitiously glued the laptop to the table).  Optical flow is a tighter concept with much less inherent ambiguity, precisely because you can ``use much more ground truth'' (all patches in both frames $f_0$ and $f_1$) when you estimate flow than you can when you estimate segments from a static image. This fundamental difference between the segments and flow concepts can be ultimately traced to the difference between mere hypotheticals and true counterfactuals.  

It is also for this reason that flow is an excellent first intermediate structure on which to kick off a virtuous cycle of structural bootstrapping.  Because it is a strong causal inference, using more information than subsequent downstream extractions, it can be estimated decently well by a $\Psi$ predictor that has so far seen only raw RGB. This situation is an example of a more general phenomenon that is critical for continual learning to occur: the stronger the causal inference, the less powerful the predictor needed to estimate it well. Thus an early (not-yet-enriched) model can be sufficiently good to support extraction of strong causal inferences (such as flow), which can then allow a model enriched with the tokens emitted by that first round of extraction to become powerful enough as a predictor to better support extraction of weaker inferences (such as segmentation).

\subsection{Limitations and Key Open Problems}
PSI has a number of limitations and leaves open a wide variety of future questions.

\textbf{Integration of Global/Object-Centric Structures.} In this work we show that spatially global structures (e.g. motion-derived object segments) can be extracted in a natural fashion using the PSI techniques. However, we only provide evidence of ``closing the cycle'' by integration gains for a local quantity (flow).  We do not yet take advantage of global structures to create ``object-centric'' predictors. It is hoped that doing so would lead to both prediction performance and control precision gains.  Ultimately, it will be useful to simultaneously integrate all the various intermediates we extract (e.g. flow, depth, segments).

\textbf{Data domains beyond vision.} The formulation of PSI is fairly general and largely independent of the specifics of the main example domain we present here, e.g. namely, natural video.  It is thus natural to imagine applying PSI to a variety of other domains, such as neural response recordings, spatial transcriptomics data, geospatial, meteorological and cosmological data, and high-resolution spatio-temporal economics datasets.  

Carrying out the first and third PSI steps in any such domain (creating an LRAS distributional predictor and integrating new token types via intermixing) seems comparatively straightforward.  However, a key unknown for many domains will be how to take the second PSI step -- namely, how to identify useful causal inferences.  In computer vision, we rely on a clear set of intermediate structures that have been discovered over the years, and which have guided our selection of useful inferences in  \S\ref{sec:structure_extraction} of this paper. How will this happen in general, for domains where intermediate structures are less well-documented?  This question motivates the next issue. 

\textbf{Automated identification of useful intermediate structures.} To create the spectrum of structures extracted in \S\ref{sec:structure_extraction}, we created each causal probe by hand. A critical forward-going question for the PSI concept will be whether it is possible to generate useful causal probes in an automated fashion, without the need for hand-coded (that is, human-intelligence-guided) discovery. A successful approach to this problem would be a kind of high-level meta-principle of data-driven structure discovery, achieving a balance between the structure-skepticism of Sutton's Bitter Lesson~\cite{Sutton2019Bitter} and the structure-optimism of developmental psychology~\cite{Spelke2007CoreKnowledge} and the Probabilistic Programming approach~\cite{Lake2015Concept}.

Realizing this idea would probably involve some kind of Domain Specific Language (DSL) that could parameterize causal inference programs, possibly exploiting ideas from the literature on automated reasoning and inference discovery in probabilistic graphical models~\cite{saad2019bayesian, glymour2019review} or language models~\cite{li2024automated}. It would also be natural if some principle of evolutionary utility could be used for causal program selection, coupled either with random or intrinsic curiosity-driven variation generation. It would also be natural to bolster this effort by deploying the principles for optimizing counterfactuals discussed in Stojanov \textit{et al.}~\cite{stojanov2025self}.

\textbf{Learned Pointer Sequences.} In this work we derive benefit from using random sequence orders in learning the $\Psi$ model, and thus provide no training signal on the pointer tokens themselves. But because some generation sequences are better than others, as shown in (\S\ref{subsec:uncertainty_management}), it would be natural and perhaps very useful to use entropy information to bootstrap the learning of pointer token sequences.

\textbf{Distillation of Extracted Structures.} A limitation of the structure extraction paradigm of PSI (and CWM before it) is that hypothetical structure extraction is fairly slow. In recent work we have illustrated a simple approach to distillation that essentially treats the structures extracted as pseudo-labels for training fast single-purpose inference models~\cite{stojanov2025self}. More generally, it is possible to formulate the distillation problem as a sparse-to-dense prediction problem in the same sequential form as the existing PSI task, but using a modified sequence order (e.g.\ RGB0-RGB1-flow'') that departs from the mix-in'' order used in PSI integration (e.g.\ ``RGB0-flow-RGB1''). Exploring these possibilities is a natural future work direction.

\textbf{Where do semantic categories fit in?} Many useful tasks that neural networks do involve explicit categorical concepts that have semantic value, such as high-level visual object categories like those that appear in the ImageNet ontology, or categorizing the actions of subjects in a video. Even when the networks are trained in a mostly self-supervised fashion (e.g.\ modern contrastive models~\cite{He2020Momentum,Chen2020Simple}), many of the loss functions for which such networks are trained produce ``cluster-y'' representations that enable simple probes (e.g. single linear layer) to recognize semantic category boundaries. Where are such concepts in PSI? It may be that PSI-learned encoders are effective at supporting category readouts in the same way as earlier self-supervised methods; but it is not obvious if they are as cluster-biased as contrastive self-supervision objectives are. Whether and to what extent contrastive techniques should be integrated with PSI is an important question.

\textbf{Long range predictions: how is memory to be integrated?}  While it is possible to scale PSI to learn decently long sequences -- and thus understand/generate videos of at least a few seconds long, and probably substantially longer -- the future of scaling both for PSI (and dedicated video generation models as well) will surely involve some kind of online-adaptive memory, akin to the hippocampal-cortical interaction~\cite{mcclelland1995there}. Exactly how best to do this to achieve long-range generation and planning remains uncertain.  

\subsection{Does PSI have anything to do with human neuroscience or cognition?} 
Frankly, we don't know yet. However, this is an obvious target for future work, both because it is an interesting class of questions in itself, and because we as a team are, in other lines of work, often engaged in comparisons of AI models to human behaviors and brains.   
Our team's strategy for creating improved cognitive and neural models has always been to look for large capability gaps between existing AI approaches and what humans do, and then seek to fill these gaps with improved algorithmic ideas~\cite{yamins2016using}. The principle behind this activity is what we have in other contexts called \emph{contravariance}: that as models are increasingly strongly constrained to solve successively more challenging intelligence goals, they are increasingly likely to converge on mechanisms like those that evolution has created in people~\cite{cao2024explanatory}. PSI can be thought of as an excursion into the \emph{terra incognita} of novel algorithm spaces, relying on the contravariance principle to eventually bring us home onto solid scientific grounding. Time will tell if this is a sound strategy. 

%% file: acknowledgements.tex
\newpage 

\section*{Acknowledgements}

This work was supported by the following awards: Simons Foundation grant SFI-AN-NC-GB-Culmination-00002986-05, National Science Foundation CAREER grant (to DLKY) 1844724, National Science Foundation Grant NCS-FR 2123963, Office of Naval Research grant N00014-20-1-2589, ONR MURI N00014-21-1-2801, ONR MURI N00014-24-1-2748, and ONR MURI N00014-22-1-2740. We also thank Stanford HAI, the Stanford Data Sciences Institute (Marlowe team), and the Google TPU Research Cloud (Zak Stone, Jonathan Caton) for their computing support.  
We thank Jay McClelland, Jim DiCarlo, Josh Tenenbaum, Shaul Druckmann, Aaron Walsman, Alyosha Efros, Jitendra Malik, Yutong Bai, Tyler Bonnen, Stephen Maturin, Justin Gardner, Jiajun Wu, Judy Fan, Tobi Gerstenberg, Konrad Koerding, Tyler Brooke-Wilson, Chaz Firestone, LA Paul, and Josh Knobe for helpful discussions during the preparation of this manuscript. We thank Greta Tuckute for the example photographs, and insightful discussions. Tian Tian Mayimin provided the scones.

\begin{center}
***

{\fontencoding{T1}\fontfamily{calligra}\selectfont 
S.D.G.}

***
\end{center}